%% file: main.tex
\documentclass{article}

\usepackage[table,xcdraw,dvipsnames]{xcolor}

\usepackage[preprint]{corl_2025} 
\newenvironment{tightlist}%
{\begin{list}{$\bullet$}{%
    \setlength{\topsep}{0in}
    \setlength{\partopsep}{0in}
    \setlength{\itemsep}{0in}
    \setlength{\parsep}{0in}
    \setlength{\leftmargin}{1.5em}
    \setlength{\rightmargin}{0in}
}
}%
{\end{list}
}

\usepackage{amssymb}

\usepackage{titletoc} 
\usepackage{amsmath}
\usepackage{booktabs}
\usepackage{graphicx}
\usepackage{comment}
\usepackage{float}
\usepackage{amsthm}
\usepackage{amssymb}
\usepackage{mathtools}
\usepackage{float}
\usepackage[normalem]{ulem}
\usepackage{wrapfig}
\usepackage{listings}
\usepackage{multicol}
\usepackage{footnote}
\usepackage{subfigure}
\usepackage{makecell}

\usepackage{listings}

\definecolor{deepblue}{rgb}{0,0,0.5}
\definecolor{deepred}{rgb}{0.6,0,0}
\definecolor{magenta}{rgb}{1.0,0,1.0}
\definecolor{deepgreen}{rgb}{0,0.5,0}
\definecolor{textblue}{rgb}{.2,.2,.7}
\definecolor{textred}{rgb}{0.54,0,0}
\definecolor{textgreen}{rgb}{0,0.43,0}
\definecolor{es-blue}{rgb}{0.1372,0.666,1}
\definecolor{lightgraybg}{RGB}{247,247,247}  

\lstdefinestyle{pythonstyle}{
    language=Python, 
    breaklines=true,
    basicstyle=\ttfamily\small,
    emphstyle=\bfseries\color{deepred}, 
    emph={forward,forward_v},         
    numbers=left,
    numberstyle=\tiny\color{gray},
    stepnumber=1,
    numbersep=12pt,
    tabsize=2,
    stringstyle=\color{textgreen},
    frame=none,                    
    columns=fullflexible,
    keepspaces=true,
    xleftmargin=\parindent,
    showstringspaces=false,
    commentstyle=\color{deepgreen},
    keywordstyle=\color{es-blue},
    backgroundcolor=\color{lightgraybg}  
}

\definecolor{promptorange}{rgb}{0.8,0.33,0}

\lstdefinestyle{promptstyle}{
    backgroundcolor=\color{lightgraybg},
    basicstyle=\ttfamily\small,
    breaklines=true,
    showstringspaces=false,
    keywordstyle=\color{es-blue},
    stringstyle=\color{deepgreen},
    commentstyle=\color{promptorange}\itshape,
    moredelim=[s][\color{gray}]{```}{```},
    moredelim=[s][\color{deepgreen}]{"""}{"""},
    morecomment=[l]{\#},
    columns=fullflexible,
    escapeinside={(*@}{@*)}, 
}

\usepackage{caption}
\usepackage[noend,]{algpseudocode}
\usepackage{fancyvrb}

\theoremstyle{definition}

\input{variables}
\input{caelan}

\definecolor{codebg}{rgb}{0.95, 0.95, 0.92}
\definecolor{codecomment}{rgb}{0.3, 0.6, 0.3}
\definecolor{codekeyword}{rgb}{0.0, 0.3, 0.7}
\definecolor{codestring}{rgb}{0.58, 0, 0.82}

\lstnewenvironment{speciallstlisting}{
    \lstset{
        language=Python, 
        backgroundcolor=\color{codebg},
        basicstyle=\ttfamily\small,
        keywordstyle=\color{codekeyword},
        commentstyle=\color{codecomment},
        stringstyle=\color{codestring},
        numbers=left,
        numberstyle=\tiny\color{codecomment},
        stepnumber=1,
        frame=single,
        framesep=5pt,
        framerule=0pt,
        rulecolor=\color{codecomment},
        tabsize=4,
        showstringspaces=false,
        breaklines=true,
        breakatwhitespace=true,
        captionpos=b
    }
}{}

\ifdefined\showcomments
  \newcommand{\caelan}[1]{\textcolor{blue}{(CG: #1)}}
  \newcommand{\nishanth}[1]{\textcolor{red}{(NK: #1)}}
  \newcommand{\willshen}[1]{\textcolor{orange}{(WS: #1)}}
  \newcommand{\todo}[1]{\textcolor{purple}{(TODO: #1)}}
\else
  \newcommand{\caelan}[1]{}
  \newcommand{\nishanth}[1]{}
  \newcommand{\willshen}[1]{}
  \newcommand{\todo}[1]{}
\fi

\newcommand{\ours}{OWL-TAMP}

\begin{document}

\title{
Open-World Task and Motion Planning via Vision-Language Model Generated Constraints
}
\author{
\textbf{Nishanth Kumar}\thanks{Work conducted partially during an internship at NVIDIA Research. Correspondence to njk@csail.mit.edu, cgarrett@nvidia.com.}\hspace{0.4em}$^{1,2}$,
\textbf{William Shen}$^{1,2}$,
\textbf{Fabio Ramos}$^{1}$, 
\textbf{Dieter Fox}$^{1,2}$, \\
\textbf{Tom\'{a}s Lozano-P\'{e}rez}$^{2}$,
\textbf{Leslie Pack Kaelbling}$^{2}$ and
\textbf{Caelan Reed Garrett}$^{1}$
\vspace{7px}
\\ 
$^1$NVIDIA Research, $^2$MIT CSAIL
}

\makeatletter

\makeatother

\maketitle


\begin{abstract}
Foundation models like Vision-Language Models (VLMs) excel at common sense vision and language tasks such as visual question answering. However, they cannot yet directly solve complex, long-horizon robot manipulation problems requiring precise continuous reasoning.
Task and Motion Planning (TAMP) systems can handle long-horizon reasoning through discrete-continuous hybrid search over parameterized skills, but rely on detailed environment models and cannot interpret novel human objectives, such as arbitrary natural language goals.
We propose integrating VLMs into TAMP systems by having them generate discrete and continuous language-parameterized {\em constraints} that enable open-world reasoning. Specifically, we use VLMs to generate discrete action ordering constraints that constrain TAMP’s search over action sequences, and continuous constraints in the form of code that augments traditional TAMP manipulation constraints. Experiments show that our approach, \ours{}, outperforms baselines relying solely on TAMP or VLMs across several long-horizon manipulation tasks specified directly in natural language.
We additionally demonstrate that \ours{} can be deployed with an off-the-shelf TAMP system to solve challenging manipulation tasks on real-world hardware. 
%
\end{abstract}


\input{introduction}

\input{related_work}

\input{problem-setting}

\input{open-world-actions}

\input{method}

\begin{figure}[t]
\centering
\includegraphics[width=\linewidth]{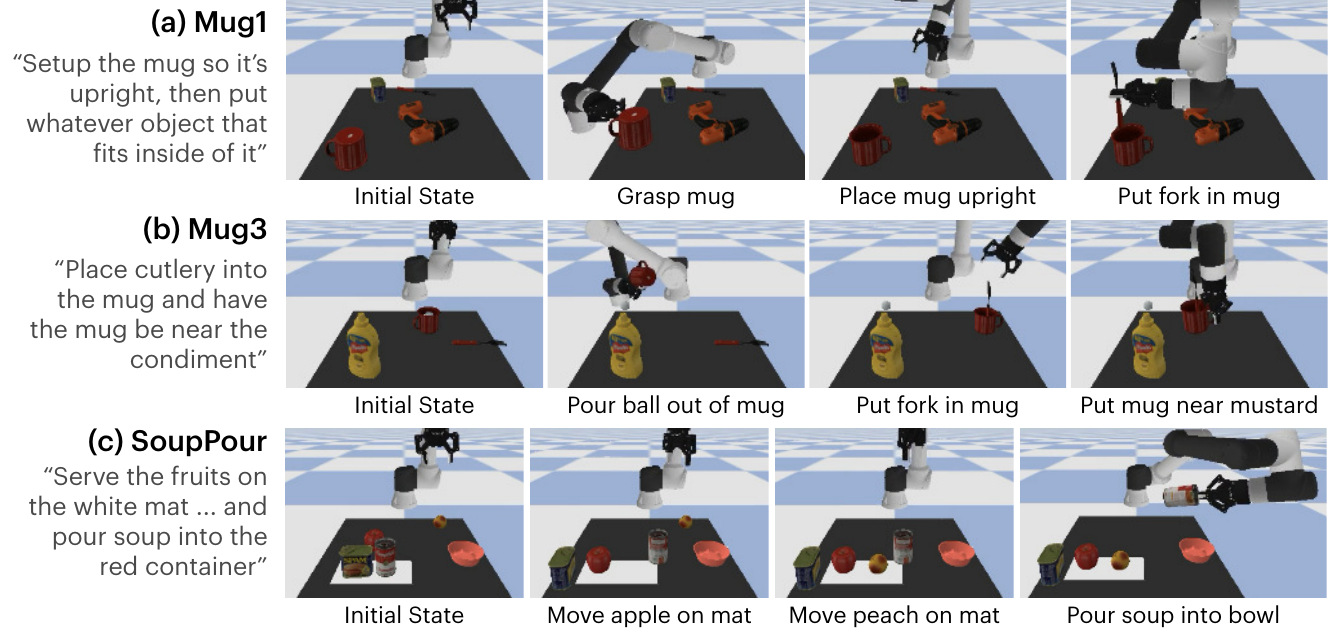}
\caption{\textbf{Simulated RAVENS-YCB Tasks.} 
\ours{}
(a) reorients the mug and infers only the fork fits inside;
(b) removes the ball from the mug before putting the fork inside, then moves it closer to the mustard;
(c) moves the spam aside to fit fruits on the mat before pouring soup into the bowl.
%
}
\label{fig:ravens-tasks}
\vspace{-15px}
\end{figure}

\input{experiments}

\input{limitations}

\input{acknowledgements}


\bibliography{references}
\clearpage
\input{appendix}



\end{document}

%% file: caelan.tex
\usepackage{algorithm}
\usepackage[noend]{algpseudocode}
\usepackage{amssymb}
\usepackage[normalem]{ulem}
\usepackage{wrapfig}
\usepackage{bm}

\newcommand{\id}[1]{{\it #1}}
\newcommand{\proc}[1]{\textsc{#1}}
\newcommand{\kw}[1]{{\bf #1}}

\newcommand{\pddl}[1]{{\texttt{#1}}} 

\usepackage{amsthm}
\theoremstyle{plain}
\theoremstyle{definition}
\theoremstyle{plain}
\theoremstyle{plain}

\usepackage{listings}


%% file: introduction.tex
\section{Introduction}
\label{sec:intro}

Solving complex, long-horizon manipulation tasks is a fundamental challenge in robotics.
Recently, there has been significant interest in solving such tasks directly from language and image input~\cite{huang2024rekep,curtis2024trustproc3ssolvinglonghorizon,code_as_policy}.
In this setting, a user provides a command like ``put away all the cutlery in the utensil holder'' and the robot must execute a sequence of actions to achieve the intended effect.
Even with pre-engineered manipulation controller primitives, such as those for moving, picking objects, and placing objects, these problems remain difficult: the robot must correctly interpret the command and sequence actions with the correct discrete arguments, such as which objects to grasp, and continuous arguments, such as grasps, stable placement poses and collision-free motions.

\begin{figure}[t]
\centering
\includegraphics[width=\linewidth]{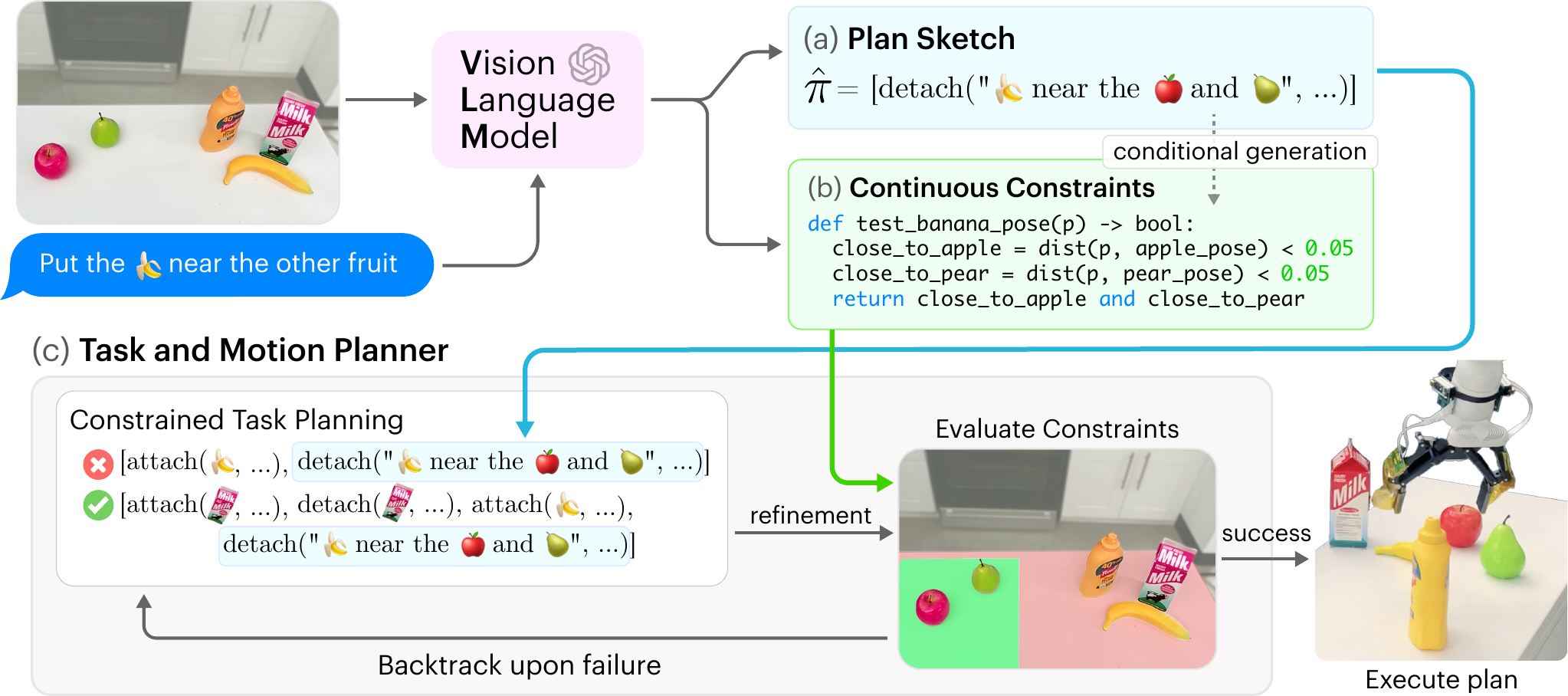}
\caption{
\textbf{\ours{} Overview}. \ours{} uses VLMs to generate task-specific open-world constraints that augment traditional robot constraints like kinematics and grasp stability.
The VLM first generates (a) plan sketches, open-world actions with natural language descriptions, and uses these to generate (b) continuous constraints in the form of code that check whether actions achieve their intended open-world predicate effects.
(c) A TAMP system consumes these constraints to solve for a plan that satisfies both traditional and VLM-generated constraints.
Here, \ours{} moves the milk carton, enabling the banana to be grasped successfully.
}
\label{fig:my-constraints}
\vspace{-15px}
\end{figure}

Consider the task in Figure~\ref{fig:my-constraints}, where the robot must ``put the banana near where the other fruits are initially.''
One approach is to prompt a Vision-Language Model (VLM)~\cite{openai2024gpt4} with the initial image, task, and available controllers~\cite{hu2023lookleapunveilingpower}.
VLMs excel at interpreting commands and leveraging common sense, and would likely suggest picking the banana before placing it. However, they struggle to reason about continuous parameter values and geometry (e.g., synthesizing collision-free grasps and placements) without task-specific fine-tuning~\cite{black2024pi0visionlanguageactionflowmodel,curtis2024trustproc3ssolvinglonghorizon}.
Here, the robot's gripper cannot safely grasp the banana without first moving obstructing objects.
By contrast, classical Task and Motion Planning (TAMP) systems explicitly reason over discrete and continuous values using symbolic predicates, operators, and geometric models~\cite{srivastava2014combined,garrett2020PDDLStream}.
A TAMP system would detect the blocked grasp and plan to move the milk carton before picking the banana.
However, TAMP systems require extensive manual modeling and cannot handle novel concepts without additional engineering.
In this case, the TAMP system lacks the concept for ``near'', and thus cannot place the banana correctly ``near'' the other fruits.

In this work, we combine the strengths of foundation models and TAMP to solve manipulation tasks specified in natural language.
Our key insight is that we can leverage the commonsense and language-grounding capabilities of VLMs to guide TAMP systems through dynamically generated \textit{constraints}.
Specifically, VLMs map language into both code expressing continuous action constraints (e.g., valid poses ``near'' other fruits) and discrete constraints \textit{plan sketches} (e.g., pick the banana, then place it ``near'' the other fruits). 
These constraints are combined with common, pre-existing TAMP constraints like collision avoidance and kinematics, and are used by a TAMP system to solve for a plan that jointly satisfies all constraints.
In our example, the TAMP system produces a correct plan that moves the milk carton to enable grasping the banana, then places it near the other fruit as dictated by the VLM-generated continuous pose constraint.
By integrating such constraints, TAMP systems become ``open-world'' -- capable of solving tasks beyond their pre-defined symbolic vocabulary (e.g., reasoning about ``near'' or ``oriented straight'' without explicit pre-defined predicates).
As a result, our approach is able to solve tasks beyond the closed-world scope of prior work that just use VLMs to accelerate discrete planning~\cite{yang2024guidinglonghorizontaskmotion}.

We propose \ours{} (\underline{O}pen-\underline{W}orld \underline{L}anguage-based \underline{TAMP}), our approach that integrates open-world concepts into TAMP via constraint generation over actions and continuous variables.
Our main contribution is to propose a clear contract --- in the form of \textit{constraints} --- for combining VLMs with generic domain-independent TAMP systems to enable zero-shot generalization to a variety of tasks.
More specifically, \ours{}: (1) generates constraints on action sequences from language to specify plan sketches; (2) generates constraints on continuous variables within these plans via code; and (3) integrates both into standard TAMP pipelines.
In simulation, \ours{} achieves higher success rates on open-world tasks than several baselines, including pure VLM and pure TAMP approaches.
We also demonstrate that it enables a real robot to solve complex, natural-language based long-horizon tasks with initial states given as raw camera images.



%% file: related_work.tex
\section{Related Work}
\label{sec:related-work}
\textbf{Task and Motion Planning (TAMP).} A standard approach for solving long-horizon robot tasks via joint discrete-continuous reasoning is TAMP~\cite{tamp_survey, bilevel-planning-blog-post}. TAMP can solve tasks expressed in natural language by translating language into symbolic goals~\cite{xie2023translating}, but assumes (1) exact translation into a pre-defined predicate set and (2) that these goals are achievable with existing symbolic operators. For instance, solving tasks like Figure~\ref{fig:my-constraints} would require a ``near'' predicate; if only an ``on top'' predicate exists, translation would fail. Expanding TAMP's capabilities thus requires manual or learned predicate invention~\cite{konidaris2018skills, silver2023predicate, liang2024visualpredicatorlearningabstractworld, athalye2024predicate}. In contrast, our approach uses pretrained VLMs without finetuning to generate constraints that expand TAMP's capabilities without task-specific demonstration or exploration data (Section~\ref{sec:experiments}).

\textbf{LLMs, VLMs and VLAs for Robotic Manipulation.} Recent work has leveraged VLMs~\cite{huang2023voxposer,huang2024copa,huang2024rekep,ma2024eurekahumanlevelrewarddesign} to enable complex short-horizon visuomotor behaviors, such as pouring water into a cup.
In contrast, we focus on longer horizon tasks composed of multiple atomic behaviors.
Additionally, to the extent these approaches do handle multi-step tasks (e.g., pick-and-place tasks), they often make assumptions or leverage heuristics specific to the particular class of tasks --- such as hardcoding the action ordering~\cite{huang2024rekep} or learning in simulation --- and are thus not zero-shot~\cite{ma2024eurekahumanlevelrewarddesign}.
More recent works attempt to solve long-horizon tasks by training Vision Language Action models (VLAs) on large amounts of robot demonstration data~\cite{black2024pi0visionlanguageactionflowmodel,nvidia2025gr00tn1openfoundation}.
However, these models require finetuning on some task and environment-specific data when adapting to novel setups and embodiments.
By contrast, our approach combines VLMs with generic, domain-independent TAMP systems to solve tasks with no requirement for additional demonstrations or learning in simulation.

\textbf{LLMs and VLMs for Robot Task Planning.} Another line of work uses foundation models for planning by sequencing discrete skills in settings where continuous parameters are either unnecessary or easily determined by heuristics~\cite{saycan, grounded_decoding, prog_prompt, inner_monologue}. Some works address continuous parameters~\cite{inner_monologue, doremi, AgiaMigimatsuEtAl2023, code_as_policy, llm3, yuan2024robopoint, duan2024manipulateanything}, but rely on foundation models to predict them directly or generate code, typically succeeding only when parameters are easily inferred from images or text (e.g., top-down grasping). A few approaches tackle complex tasks requiring precise discrete and continuous parameters~\cite{replan, tamp_llm, chen2024autotamp, curtis2024trustproc3ssolvinglonghorizon}, but struggle with discrete-level \textit{backtracking} over action sequences or long-horizon tasks where explicit search is preferable~\cite{kambhampati2024can}. Our method instead combines VLMs with off-the-shelf TAMP systems to handle long-horizons and more complex constraints.
Recent work~\cite{yang2024guidinglonghorizontaskmotion} leverages VLMs to generate subgoals for a TAMP system to solve sequentially, but does not generate any continuous constraints.
Concurrent work~\cite{guo2024castlconstraintsspecificationsllm} also defines a constraint-based contract between LLMs and TAMP, but only handles discrete constraints and requires a custom TAMP solver.

Overall, while there are prior works that explore the general idea of combining foundation models with search-based planning, our approach is the {\em first} to directly integrate with full off-the-shelf TAMP systems.
In \ours{}, the foundation model is not simply accelerating search but rather writing open-world continuous and discrete constraints to augment the model that search is being performed on.

%% file: problem-setting.tex
\section{Problem Formulation} 
\label{sec:formulation}
We aim to solve tasks specified in natural language, given an initial scene observation, a set of objects, a library of manipulation controllers, and a pre-existing Task and Motion Planning (TAMP) system with symbolic components.
We define a task as a tuple \(\langle s_0, \mathcal{A}, \mathcal{I}, L\rangle\), where \(s_0\) is the initial state of the robot and a set of named objects (identified via object detection and segmentation) in the environment (e.g., joint positions, object poses obtained from vision-based pose estimation), \(\mathcal{A}\) is a set of \textit{parameterized actions}, \(\mathcal{I}\) is a set of initial image observations of the scene, and \(L\) is the natural language instruction (e.g., ``put the banana near the other fruit" in Figure~\ref{fig:my-constraints}).

Each parameterized action in \(\mathcal{A}\) consists of a list of typed parameters, which may be discrete (e.g., object identifiers) or continuous (e.g., object poses or robot configurations), and a low-level controller (e.g., a joint-trajectory controller) that executes the action according to the parameter values.
%
For example, consider the typed parameters of an \pddl{attach} action:
\pddl{attach}(\(o{:}\;\pddl{obj}, p{:}\;\pddl{pose}, g{:}\;\pddl{grasp}, q{:}\;\pddl{conf}\)).
%
Here, \(o\) is a discrete object identifier, \(p \in \text{SE}(3)\) is an approach pose near object \(o\), \(g \in \text{SE}(3)\) is a grasp pose where the robot closes its fingers, and \(q \in \mathbb{R}^d\) is the corresponding robot joint configuration used to execute grasp \(g\) on \(o\).
Specifying concrete values for all parameters \textit{grounds} the action, yielding a fully executable manipulation controller.

The robot's objective is to generate and execute a \textit{plan}: a sequence of grounded actions which, when applied from the initial state \(s_0\), results in a final state \(s_*\) that satisfies the goal described by \(L\).

%
%

\subsection{Task and Motion Planning}
\label{subsec:TAMP-background}

Solving the tasks described above is challenging because the robot must not only select a valid sequence of actions, but also ground each action with precise parameters that achieve the goal described by \(L\).
We are interested in leveraging a TAMP system to help address these challenges; we thus assume access to a model-based TAMP system that is fixed and shared across all tasks.

\begin{figure*}[t]
\vspace{0.25cm} 
\includegraphics[width=\linewidth]{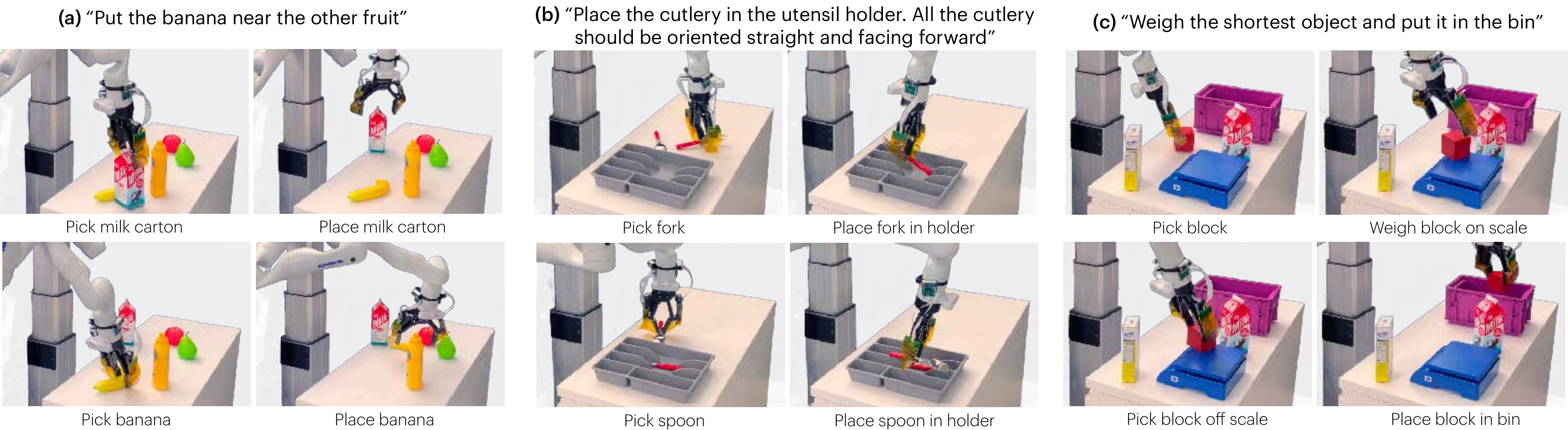}
\caption{
\textbf{Real-World Tasks.}
%
\ours{}
(a) reasons to move the milk obstructing the banana, then places the banana near the other fruit (Figure~\ref{fig:my-constraints}).
%
%
(b) plans placements satisfying both the language goal and collision constraints.
(c) identifies the block as shortest, and 
weighs it on the scale before disposing.
}
\label{fig:my-rummy}
\vspace{-15px}
\end{figure*}

Within TAMP, we model states and actions using a Planning Domain Definition Language (PDDL)-style~\cite{mcdermott1998pddl} factored action language, where states are sets of true-valued literals, i.e., \textit{predicates} applied to specific parameters.
For example, the initial state \(s_0\) in Figure~\ref{fig:my-constraints} includes [\pddl{AtConf}(\(q_0\)), \pddl{HandEmpty}(), \pddl{AtPose}(\pddl{banana}, \(p_0\)), \pddl{On}(\pddl{banana}, \pddl{table}),\;\dots], indicating that the robot is at configuration \(q_0\), its gripper is empty, the banana is at placement pose \(p_0\) and on the table.

Each parameterized action \(a\) in \(\mathcal{A}\), defined as part of the task, is extended in the TAMP system with a set of constraints (\textbf{con}) that the parameters must satisfy, preconditions (\textbf{pre}) that must hold before applying the action, and effects (\textbf{eff}) that describe how the state changes after executing action \(a\).
The actions \pddl{move} and \pddl{attach} model the robot moving between two configurations and attaching an object to itself via grasping, respectively.

\vspace{-5px}
\begin{minipage}{0.4\textwidth}
\begin{footnotesize}
\begin{lstlisting}
move|$(q_1: \text{conf},\; q_2: \text{conf},\; \tau: \text{traj})$|
 |\kw{con}:| [Motion|$(q_1, \tau, q_2)$|]
 |\kw{pre}:| [AtConf|$(q_1)$|]
 |\kw{eff}:| [AtConf|$(q_2)$|, |$\neg$|AtConf|$(q_1)$|]
attach|$(o: \text{obj},\; p: \text{pose},\; g: \text{grasp},\; q: \text{conf})$|
 |\kw{con}:| [Kin|$(q, o, g, p)$|]
 |\kw{pre}:| [AtPose|$(o, p)$|, HandEmpty|$()$|, AtConf|$(q)$|]
 |\kw{eff}:| [AtGrasp|$(o, g)$|, |$\neg$|AtPose|$(o, p)$|, |$\neg$|HandEmpty|$()$|]
\end{lstlisting}
\end{footnotesize}
\end{minipage}
\vspace{-8pt}


The
\(\pddl{Motion}(q_1, \tau, q_2)\)
constraint checks that \(\tau\) is a valid trajectory connecting configurations \(q_1\) and \(q_2\), while
\(\pddl{Kin}(q, o, g, p)\)
checks whether configuration \(q\) satisfies a kinematic constraint for grasping object \(o\) at  placement pose \(p\) with grasp \(g\). The \(\pddl{AtGrasp}(o, g)\) predicate indicates that object \(o\) is currently grasped with grasp pose \(g\).
Additionally, for some of these constraints, we have standard robotics samplers that generate continuous values that satisfy these constraints, namely an inverse kinematics solver for \pddl{Kin} and motion planner for \pddl{Motion}.
The challenge in planning is \textit{refinement}: generating input parameter values for these actions (e.g. the pose \textit{p} and grasp \textit{g} for \pddl{attach}).

The goal in TAMP is typically specified by a symbolic goal \(G\), a set of literals that must hold in the final state.
The planner searches for a sequence of grounded actions that results in a state \(s_*\) where all literals in \(G\) evaluate to true.
This search is conducted within a 3D model of the environment and objects: in real-world environments, this is constructed from the robot's initial sensory observations such as point clouds.
A key feature of TAMP is that the planner is capable of not just considering a single action sequence to achieve the goal (e.g., directly \pddl{attach}-ing then \pddl{detach}-ing the banana in Figure~\ref{fig:my-constraints}c), but rather exploring multiple sequences (via \textit{backtracking} over an abstract plan) across multiple possible action sequences (e.g., first moving the milk carton out of the way) until it finds one that  accomplishes the task.

%% file: open-world-actions.tex
\section{Open-World Actions and Predicates}
\label{sec:open-world}

Traditional TAMP systems rely on a fixed set of symbolic predicates and operators, which limits their applicability to tasks that can be fully specified within this  vocabulary.
However, many open-ended tasks in unstructured environments involve object semantics, spatial relationships, or context-dependent goals that are not explicitly encoded in the domain definition.
To address this, we propose a new abstraction that extends TAMP representations through \textit{open-world actions}, which augment standard TAMP actions with natural language parameters, and \textit{open-world predicates}, whose behavior is dynamically grounded via natural language.

\textbf{Open-World Actions.} An open-world action extends a parameterized TAMP action (Section~\ref{subsec:TAMP-background}) with an additional discrete argument: a natural language \pddl{description} \(d\) as a string.
%
This description specifies task-specific intent that constrains the set of valid continuous parameter values, such as object poses or robot configurations.
Open-world actions enable novel behaviors that go beyond what can be expressed using the fixed symbolic vocabulary of a traditional TAMP system.
For example, in Figure~\ref{fig:my-constraints}, the \pddl{detach} action is invoked with the description \pddl{detach}(``place the banana near the apple and pear",\; \dots), which narrows the feasible placement region from the entire table to a localized region, as visualized by the colored regions.
A different description, such as ``place the banana atop the orange" would restrict the final pose of the banana to be stably resting on the orange.

\textbf{Open-World Predicates.} To enforce that the intent expressed by the language description \(d\) is achieved, we introduce open-world predicates, which are appended to the constraint list (\textbf{con}) of an open-world action.
Traditional predicates rely on hand-engineered, fixed classifiers to evaluate whether a \textit{literal} --- a predicate with fully specified parameter values --- is considered true or false.
In contrast, an open-world predicate uses a classifier that is generated dynamically at planning time, conditioned on the description \(d\) and other parameters (Figure~\ref{fig:my-constraints}b).
We assume that these predicates take in \(d\) as a parameter, along with any other objects and continuous parameters, and output a boolean value indicating that the parameters are set such that \(d\) is achieved.
We specifically use a VLM to generate this classifier function in the form of Python code (Section~\ref{subsec:vlm-continuous-constraint-gen}).
The resulting classifiers act as constraints during planning, restricting the set of valid continuous parameters (e.g., object poses) according to whether they satisfy the task-specific intent described by \(d\).
%

\textbf{Example.} We make the \pddl{detach} action open-world by modeling the description \(d\) as a parameter and including the open-world predicate \pddl{VLMConstraint} in its constraints:


\vspace{-2pt}
\begin{footnotesize}
\begin{lstlisting}
detach|$(\underline{d{:}\; \text{description}},\; o{:}\; \text{obj},\; g{:}\; \text{grasp},\; p{:}\; \text{pose},\; q{:}\; \text{conf})$|
  |\kw{con}:| [Kin|$(q, o, g, p)$|, |\underline{VLMConstraint$(d, o, p)$}|]
  |\kw{pre}:| [AtGrasp|$(o, g)$|, AtConf|$(q)$|,
        |$\neg \exists o', p'.\; \pddl{AtPose}(o', p') \wedge \pddl{Collision}(o, p, o', p')$|]
  |\kw{eff}:| [AtPose|$(o, p)$|, |$\neg$|AtGrasp|$(o, g)$|, HandEmpty|$()$|]
\end{lstlisting}
\end{footnotesize}

Let the open-world predicate
$\pddl{VLMConstraint}(d, o, p)$
be true if placing object \(o\) at placement pose \(p\) satisfies description \(d\).
This allows the \pddl{detach} action to jointly respect both task-specific constraints specified by \(d\) and engineered constraints such as kinematic feasibility (\pddl{Kin}).
In Figure~\ref{fig:my-constraints}c, the TAMP planner rejection samples the set of reachable poses (generated by samplers for \pddl{Kin}) against \pddl{VLMConstraint}, implemented by the VLM-generated classifier function \pddl{test\_banana\_pose}, to yield a collision-free banana placement that is near the other fruit.

The \pddl{detach} action serves as a simple illustrative example that operates over a single object.
Open-world actions can more generally use any number of objects and parameters, and model a variety of other interaction types.
The following \pddl{transport} action models moving an object $o$ through pose waypoints $p_1$ and $p_2$ to then interact with object $o'$. The waypoints $p_1$ and $p_2$ are determined by a VLM constraint \pddl{VLMConstraint2} and can, for example, represent both the start and end of a pouring motion involving a cup (e.g. in the \textit{SoupPour} task in Section~\ref{sec:experiments}) into a bowl $o'$ as well as filling a cup $o$ from a faucet $o'$.
The key takeaway is that open-world actions enable planning with both manually engineered TAMP constraints and dynamically generated, task-specific constraints grounded in natural language, expressed as open-world predicates.

\begin{footnotesize}

\begin{lstlisting}
transport|$(\underline{d{:}\; \text{description}},\; o{:}\; \text{obj},\; g{:}\; \text{grasp},$|
    |$p_1{:}\; \text{pose},\; p_2{:}\; \text{pose},\; o'{:}\; \text{obj}, q_1{:}\; \text{conf},\; q_2{:}\; \text{conf})$|
  |\kw{con}:| [Kin|$(q_1, o, g, p_1)$|, Kin|$(q_2, o, g, p_2)$|,
        |\underline{VLMConstraint2$(d, o, p_1, p_2, o')$}|]
  |\kw{pre}:| [AtGrasp|$(o, g)$|, AtConf|$(q_1)$|]
  |\kw{eff}:| [AtConf|$(q_2)$|, |$\neg$|AtConf|$(q_1)$|]
\end{lstlisting}
\end{footnotesize}



%% file: method.tex
\section{Open-World Language-Based TAMP} 
\label{sec:tamp-with-open-world-concepts}

Given a natural language instruction, images of the initial scene, a set of objects segmented within a scene, a TAMP system, and a collection of open-world actions and predicates, \ours{} produces a valid plan that satisfies both manually engineered symbolic constraints
and dynamically generated constraints from a VLM.
Our approach operates in three stages:
(1) it prompts a VLM to generate a discrete \textit{plan sketch} of open-world actions relevant to the task (Section~\ref{subsec:action-ordering});
%
(2) it grounds open-world predicates in the sketch into executable \textit{continuous constraints} via VLM-generated classifier functions (Section~\ref{subsec:vlm-continuous-constraint-gen});
and (3) it solves the resulting TAMP problem using an off-the-shelf planner (Section~\ref{subsec:tamp-with-vlm-cons}).

\subsection{Generating Open-World Action Ordering Constraints}
\label{subsec:action-ordering}

Given the language instruction \(L\) and initial images \(\mathcal{I}\) (Section~\ref{sec:formulation}), we first prompt a VLM to produce a \textit{plan sketch} \(\hat{\pi}\): a partial sequence of open-world actions with discrete object arguments and natural language descriptions that reflect how to achieve the task specified by \(L\).
%
This sketch encodes a high-level plan structure, specifying a minimal set of actions and their order for a plan to be valid.
%
In the banana task (Figure~\ref{fig:my-constraints}), the sketch could be: %
\\\indent$\hat{\pi}$ = [\pddl{detach}(\text{``place the banana near the apple and pear"},\;
\pddl{banana},\;\dots)].
\\In the weigh and dispose shortest object task (Figure~\ref{fig:my-rummy}c), the sketch might be: %
\\\indent$\hat{\pi} {=} $[\pddl{detach}(\text{``weigh block on scale"},\;\pddl{block},\;\dots),\; \pddl{detach}(\text{``place block in bin"},\;\pddl{block},\;\dots)],
\\allowing us to \pddl{detach} the same object on different support surfaces.
However, VLM-generated sketches are not guaranteed to be complete -- they may omit necessary actions (e.g., forgetting to \pddl{attach} before \pddl{detach}) or propose actions that are infeasible due to low-level constraints such as collisions or kinematic feasibility.
In the banana task (Figure~\ref{fig:my-constraints}), the VLM may omit a required \pddl{attach}(\pddl{banana}, \dots) before \pddl{detach}, or fail to consider that the milk carton obstructs all grasps on the banana.

To address these issues, we treat the plan sketch as a set of discrete action-ordering constraints.
Specifically, we constrain the TAMP planner to generate plans that include \(\hat{\pi}\) as a subsequence -- that is, the specified actions must appear in the given order, though additional actions may be inserted before, between, or after them to ensure the plan is feasible.

We generate plan sketches using a few-shot prompting approach with a VLM with no fine-tuning, and describe how they are symbolically encoded into a TAMP system in Appendix~\ref{subsec:appendix-discrete-constraint-generation}.

To enable a VLM to generate a plan sketch for a given task, we associate a natural language description of each available action with that particular action.
Although we could directly prompt a VLM for relevant actions and goals, without a list of candidates, the VLM is likely to be syntactically and semantically inaccurate.
Instead, we first ground the set of {\em reachable} actions $A$ and literals $\Lambda$ (i.e., grounded predicates) available to the TAMP system before prompting the VLM to return values in these sets.
We use {\em relaxed planning}~\cite{bonet1999planning} from the initial state $s_0$ to simultaneously ground~\cite{HoffmannN01,edelkamp2000exhibiting} and explore the sets of reachable actions $A$ and literals $\Lambda$.
We use placeholders to instantiate continuous parameters, such as {\em optimistic} values~\cite{garrett2021sampling}, to ensure a finite set of actions are instantiated.
Similarly, we use placeholders for \pddl{description} parameters.

\begin{algorithm}[th] 
\begin{small}
  \caption{VLM Task Reasoning}
  \label{alg:planning}
  \begin{algorithmic}[1] 
    \Procedure{vlm-task-reasoning}{$s_0, {\cal A}, {\cal I}, L$}
        \State $A \gets \proc{ground-actions}(s_0, {\cal A})$
        \State $\Lambda \gets s_0 \cup \{l \mid a \in A.\; l \in a.\id{eff}\}$
        \State $[a_1, ..., a_n, l_1, .., l_k] \gets \proc{query-vlm}(\text{``What partial plan} \newline \text{using actions } \{A\} \text{ for goal literals } \{\Lambda\} \text{ achieves goal } \{L\}\text{?''})$
        \For{$i \in [1, n-1]$}
            \State $a_{i}.\id{eff} \gets a_{i}.\id{eff} \cup \{\pddl{Executed}(i)\}$
            \State $a_{i+1}.\id{pre} \gets a_{i+1}.\id{pre} \cup \{\pddl{Executed}(i)\}$
        \EndFor
        \State $a_{n}.\id{eff} \gets a_{n}.\id{eff} \cup \{\pddl{Executed}(n)\}$
        \State $G \gets \{l_1, .., l_k\}$
        \State \Return $\proc{solve-tamp}(s_0, A, G \cup \{\pddl{Executed}(n)\})$ 
    \EndProcedure
\end{algorithmic}
\end{small}
\end{algorithm}

Algorithm~\ref{alg:planning} presents the VLM partial planning pseudocode.
It takes in a TAMP problem $\langle s_0, {\cal A}, \mathcal{I}, L \rangle$, where $L$ is a text goal description and \(\mathcal{I}\) is a set of initial image observations of the scene.
It first grounds the set of actions $A$ reachable from $s_0$ using \proc{ground-actions}.
Then, it accumulates the set of reachable literals $\Lambda$ by taking the effects of all actions $A$.
Then, it prompts \proc{query-vlm} for a partial plan $[a_1, ..., a_n, l_1, ..., l_k]$ using actions $a_i \in A$ and goal literals $l_j \in \Lambda$ that achieve $L$.
Here, the literals $l_j \in \Lambda$ are intended to be a translation of the goal into atoms, while the sequence of actions is a way of achieving that goal.
Importantly, we have the VLM fill in the description parameter $d$ for each of these actions.
We transform the original problem to so that solutions admit the partial plan as a subsequence.
Specifically, we create a predicate \proc{Executed} that models whether the $i$th action in the plan was executed and add $\proc{Executed}$ to the effects of action $a_i$ and the preconditions of action $a_{i+1}$.
Finally, we make the planning goal $G = \{l_i, ..., l_m\} \subseteq L$ and $\proc{Executed}(n)$.

\begin{figure*}[t]
\centering
\vspace{0.25cm} 
\begin{minipage}{\textwidth}
\centering
\setlength{\tabcolsep}{2pt}
\begin{scriptsize}
\resizebox{\textwidth}{!}{%
\begin{tabular}{lrrrrrrrrrrr}
\toprule
\multicolumn{1}{c}{} & \multicolumn{10}{c}{\textbf{Tasks}} \\
\cmidrule(lr){2-11}
\textbf{Method} & {Berry1} & {Citrus} & {Berry2} & {BerryCook} & {FruitSort} & {Coffee} & {Mug1} & {Mug2} & {Mug3} & {SoupPour} & \textbf{Overall} \\
\midrule
CaP
& 0\% $\pm$ 0.0  & 0\% $\pm$ 0.0  & 0\% $\pm$ 0.0  & 0\% $\pm$ 0.0  & 0\% $\pm$ 0.0
& 0\% $\pm$ 0.0  & 0\% $\pm$ 0.0  & 0\% $\pm$ 0.0  & 0\% $\pm$ 0.0  & 0\% $\pm$ 0.0
& 0\% $\pm$ 0.0 \\
CaP-sample
& \textbf{100\% $\pm$ 0.0} & 20\% $\pm$ 12.6 & 20\% $\pm$ 12.6 & 0\% $\pm$ 0.0 & 0\% $\pm$ 0.0
& 0\% $\pm$ 0.0 & 0\% $\pm$ 0.0 & 0\% $\pm$ 0.0 & 0\% $\pm$ 0.0 & 0\% $\pm$ 0.0
& 14\% $\pm$ 3.4 \\
\midrule
Direct trans.
& \textbf{100\% $\pm$ 0.0} & \textbf{100\% $\pm$ 0.0} & \textbf{100\% $\pm$ 0.0}
& 0\% $\pm$ 0.0 & 0\% $\pm$ 0.0 & 0\% $\pm$ 0.0 & 0\% $\pm$ 0.0 & 0\% $\pm$ 0.0
& 20\% $\pm$ 12.6 & 0\% $\pm$ 0.0
& 32\% $\pm$ 4.7 \\
No sample
& 0\% $\pm$ 0.0  & 0\% $\pm$ 0.0  & 0\% $\pm$ 0.0  & 0\% $\pm$ 0.0  & 0\% $\pm$ 0.0
& 0\% $\pm$ 0.0  & 0\% $\pm$ 0.0  & 0\% $\pm$ 0.0  & 0\% $\pm$ 0.0  & 0\% $\pm$ 0.0
& 0\% $\pm$ 0.0 \\
No disc.
& \textbf{100\% $\pm$ 0.0} & \textbf{100\% $\pm$ 0.0} & \textbf{100\% $\pm$ 0.0}
& 0\% $\pm$ 0.0 & 0\% $\pm$ 0.0 & 0\% $\pm$ 0.0 & 0\% $\pm$ 0.0 & 0\% $\pm$ 0.0
& 20\% $\pm$ 12.6 & 0\% $\pm$ 0.0
& 32\% $\pm$ 4.7 \\
No cont.
& \textbf{100\% $\pm$ 0.0} & \textbf{100\% $\pm$ 0.0} & \textbf{100\% $\pm$ 0.0} & \textbf{100\% $\pm$ 0.0}
& 10\% $\pm$ 9.5 & 60\% $\pm$ 15.5 & 70\% $\pm$ 14.5 & 0\% $\pm$ 0.0 & 0\% $\pm$ 0.0
& 20\% $\pm$ 12.6
& 56\% $\pm$ 5.0 \\
No back.
& \textbf{100\% $\pm$ 0.0} & \textbf{90\% $\pm$ 9.5} & \textbf{100\% $\pm$ 0.0}
& 60\% $\pm$ 15.5 & 80\% $\pm$ 12.6 & \textbf{100\% $\pm$ 0.0}
& 40\% $\pm$ 15.5 & 30\% $\pm$ 14.5 & 0\% $\pm$ 0.0 & 0\% $\pm$ 0.0
& 60\% $\pm$ 4.9 \\
\midrule
OWL-TAMP
& \textbf{100\% $\pm$ 0.0} & \textbf{100\% $\pm$ 0.0} & \textbf{100\% $\pm$ 0.0}
& 60\% $\pm$ 15.5 & \textbf{100\% $\pm$ 0.0} & \textbf{100\% $\pm$ 0.0}
& \textbf{100\% $\pm$ 0.0} & \textbf{100\% $\pm$ 0.0}
& \textbf{70\% $\pm$ 14.5} & \textbf{90\% $\pm$ 9.5}
& \textbf{92\% $\pm$ 2.7} \\
\bottomrule
\end{tabular}
}
\end{scriptsize}

\captionof{table}{
\textbf{Success rates on all tasks.}
Entries report the mean success rate over 10 random seeds and the binomial standard error in percentage points.
We bold methods that are not significantly worse than the best method in that column under a one-tailed two-proportion $z$-test ($\alpha{=}0.05$).
The final column pools successes across all 10 tasks ($n{=}100$).
}
\label{tab:method_env_results}
\vspace{-10px}
\end{minipage}

%
\end{figure*}

\subsection{Grounding Open-World Predicates with a VLM}
\label{subsec:vlm-continuous-constraint-gen}

After obtaining a plan sketch from a VLM, we ground each open-world predicate into a concrete constraint that can be evaluated during planning.
These predicates, which appear in the open-world actions within the plan sketch \(\hat{\pi}\), are grounded by dynamically generating a \textit{classifier function} for each predicate separately once the predicate is instantiated with a specific natural language description.
The resulting classifier operates over continuous-valued arguments and returns a boolean indicating whether a given literal (i.e., a predicate with fully specified arguments), is considered true or false.
We refer to these classifier functions as \textit{continuous constraints}.

Recall that the predicate
\(\pddl{VLMConstraint}(d, o, p)\)
in \pddl{detach} is used to constrain the placement of object \(o\) based on the language description \(d\).
%
When instantiated with \(d = \text{``place the banana near the apple and the pear"}\) and \(o = \pddl{banana}\), \ours{} prompts a VLM to generate a Python function \texttt{test\_banana\_pose(p)} that returns true if pose \(p\) lies near the apple and pear on the table (Figure~\ref{fig:my-constraints}b).
During planning, the TAMP system evaluates this constraint by applying the function to candidate placement poses of the banana.
We follow prior work~\cite{code_as_policy,curtis2024trustproc3ssolvinglonghorizon} and prompt a VLM separately for each predicate and description using a few-shot template and a codebook of basic helper functions (e.g., for computing bounding boxes or checking collisions between two objects), which the VLM can compose to construct more complex constraints (Appendix~\ref{subsec:appendix-continuous-constraint-helpers}).



\subsection{Planning with VLM-Generated Constraints}
\label{subsec:tamp-with-vlm-cons}

Given the discrete plan sketch and continuous constraints generated by the VLM, we can now solve the TAMP problem via an off-the-shelf planning algorithm, such as SeSaME~\cite{bilevel-planning-blog-post} or PDDLStream~\cite{garrett2020PDDLStream} (see Appendix~\ref{subsec:appendix-tamp-details}).
%
The planner searches for a valid sequence of discrete actions, constrained by the plan sketch, and solves for continuous parameter assignments satisfying both manually engineered and VLM-generated constraints.
In the banana task (Figure~\ref{fig:my-constraints}), the planner first attempts to \pddl{attach} and \pddl{detach} the banana directly, but fails to find collision-free grasps due to the obstructing milk carton.
It then backtracks and finds a feasible plan that first moves the milk carton before grasping and placing the banana at a pose that satisfies \pddl{VLMConstraint} and is reachable.


%% file: experiments.tex
\section{Experiments and System Demonstration}
\label{sec:experiments}

\textbf{Tasks.} We evaluate on ten open-world manipulation tasks, spanning a variety of skills including  obstacle removal, sorting, reorientation, insertion, and pouring.
All tasks implemented in the RAVENS-YCB environment on a UR5 arm, adapted from prior work~\cite{curtis2024trustproc3ssolvinglonghorizon,zeng2020transporter} (see Appendix~\ref{appendix:ravens-details}).
\begin{tightlist}
    \item \textit{Berry1}: ``put the strawberry onto the light-grey region at the center of the table''.
    \item \textit{Citrus}: ``pack the citrus fruit onto the plate''.
    \item \textit{Berry2}: Same as \textit{Berry1}, but requires moving an obstacle out of the way of the light-grey region.
    \item \textit{BerryCook}: ``Cook the strawberry by putting it in the pan, then finally simply place it in the bowl. The strawberry should only be in the bowl at the end!''.
    \item \textit{FruitSort}: ``Put all the fruit to the left of the line bisecting the table''.
    \item \textit{Coffee}: ``I want to pour some coffee into the cup; can you set up the cup on the table so I can do this properly?'' (requires reorienting the cup so that it is placed `right-side-up' on the table).
    \item \textit{Mug1}: ``Setup the mug so it's upright, then put whatever object that fits inside of it''.
    \item \textit{Mug2}: ``Place cutlery inside the mug and then place the mug itself on the table near the condiment'' (the mug's opening is obstructed by a large orange, which must be moved out of the way).
    \item \textit{Mug3}: Same as \textit{CutleryInMug2}, except that in the initial state, the mug is unobstructed by an orange but instead contains a ball, which must be removed by 'pouring' it out.
    \item \textit{SoupPour}: ``Serve the fruits on the white mat (make sure the peach is to the right of the apple'' and pour soup into the red container''. The white mat is originally obstructed by the soup can as well as a spam can, which must be moved out of the way to successfully place the fruits.
\end{tightlist}

\textbf{Approaches.} We list the various approaches we compare to \ours{} across various tasks.
\begin{tightlist}
\vspace{-2.5px}
\item \textit{CaP:} Application of Code as Policies~\cite{code_as_policy}, using a VLM to generate plan code directly from initial states. Few-shot examples and helper functions match \ours{}. 
\item \textit{Direct trans.:} Prompting an LLM to translate natural language goals to symbolic goals (see Section~\ref{subsec:TAMP-background}), then solving with TAMP. This baseline is inspired by~\cite{llm_plus_p,xie2023translating}
\item \textit{CaP-sample:} CaP variant inspired by~\cite{curtis2024trustproc3ssolvinglonghorizon}, where the VLM samples multiple continuous values for skills 
and selects the first set that satisfies robot constraints.
\item \textit{No sample:} A version of \ours{} allowing only one continuous sample per skill.
\item \textit{No cont.:} Ablation of \ours{} removing continuous constraint generation (continuous values come purely from TAMP). This is inspired by~\cite{yang2024guidinglonghorizontaskmotion}.  
\item \textit{No disc.:} Ablation of \ours{} removing action-ordering constraints.
\item \textit{No backtrack:} Ablation of \ours{} where only the first plan skeleton is refined.
\end{tightlist}

\textbf{Experimental Setup.}
We use GPT-4o~\cite{openai2024gpt4} as the VLM across all methods.
Results are averaged over 10 seeds per task, with initial object poses randomized and language goals fixed.
CaP-sample uses a 2500-sample budget per task; other methods (except CaP and No-sample, which use 1-sample) have a 500-sample budget per action.
Backtracking methods (all \ours{} variants except No-backtrack) may attempt to refine up to 5 plan skeletons.
Success is measured by task completion without constraint violation, as judged by manually written detectors; soundness rate (false positive rate) is also reported.
Additional metrics (wall-clock time, number of skeletons, and model queries) are in Appendix~\ref{appendix:additional-results}.
All methods use the same standard search-then-sample TAMP system~\cite{srivastava2014combined,bilevel-planning-blog-post,silver2023predicate,kumar2023learning}, which we detail in Appendix~\ref{subsec:appendix-tamp-details}.

\subsection{Results and Analysis}
\label{subsec:sim-results}

Table~\ref{tab:method_env_results} reports success rates. \ours{} achieves the highest success rate on 9 of 10 tasks, outperforming all baselines.
Direct trans.\@ succeeds on the 3 simplest tasks, where language goals directly map to existing TAMP predicates, but fails on harder tasks requiring language-conditioned constraint generation.
CaP and No sample find correct discrete plans but fail to select continuous parameters satisfying constraints.
CaP-sample improves with sampling but struggles when satisfying constraints requires coordinated reasoning across discrete actions.
Ablations disabling discrete (No disc.\@) or continuous (No cont.\@) constraint generation succeed only on easier tasks, while No backtrack struggles on longer-horizon problems requiring obstacle removal.
By generating both discrete and continuous constraints, \ours{} solves all tasks without any task-specific modifications.

Figure~\ref{fig:soundness_results} reports soundness rates.
Direct trans.\@ and No cont.\@ often declare success despite violating the true task intent (a false positive).
\ours{} produces almost no false positives --- only a single case across all tasks --- demonstrating precise grounding of language into constraints.
CaP-sample has a high soundness only because it simply fails on most tasks (14\% success rate) and declares it fails. 
%

Failures for \ours{} occurred in BerryCook, Mug3, and SoupPour: in BerryCook, the VLM generated an incorrect continuous constraint requiring the strawberry to be inside the bowl and pan simultaneously; in Mug3, the sample budget was exhausted despite correct generated constraints; and in SoupPour, the VLM generated an unsatisfiable plan sketch.
By far the most common failure mode was that observed in BerryCook: the system generated infeasible continuous constraints that were impossible to satisfy.
While we analyzed \ours{} as a single-shot planner, when deployed in a replanning policy it can use observations from execution or human feedback to augment the language instruction and adapt its behavior after failures.

\subsection{Real-World System Deployment}
\label{subsec:real-world-deployment}
In order to demonstrate our system's applicability to open-world tasks in the real-world, we deployed \ours{} on a dual-arm robot (Kinova Gen3 arms with a pan-tilt head camera) to perform 19 natural language manipulation tasks, listed below (Figure~\ref{fig:my-rummy}, full list in Appendix~\ref{appendix:additional-experiment-details}).
Please see our supplement for video of the robot's execution.

\begin{footnotesize}
\begin{tightlist}
    \item ``Put the orange and apple on the plate.''
    \item ``Place the strawberry and lime each in the bin that matches their color.'',
    \item ``Stack the blocks into a tower by increasing hue.''
    \item ``Put the apple left of the plate and the orange on the table surface behind of the plate.''
    \item ``Put the orange on the far right of the table and the apple on the far left''.
    \item ``Put the orange where the apple is initially''.
    \item ``Clean the plate'' (a sponge is among several other objects present on a tabletop, and the robot must put the sponge atop the plate)
    \item ``Throw away anything not vegan in the purple bin'' (objects on the table include a milk carton, apple, spam can, and water bottle).
    \item ``Put the green block between the blue and red ones''
    \item ``Put the blue block onto the plate'' (the plate is packed with distractors and the robot must make a tightly-constrained placement).
    \item ``Setup the cutlery for someone to eat a meal from the plate. All the cutlery should be close to and lined-up with the plate, and should be oriented so each is straight and facing forwards, though you should pick which side of the plate each of the items are on'' (there are two pieces of fruit, and two similarly-colored blocks that must be disambiguated).
    \item ``Fit one of the fruit in the cup'' (only one of 4 available fruits is small enough to fit in the provided mug)
    \item ``Put the brownie ingredients in front of the pan'' (only 2 of the available items are related to brownies)
    \item ``Place the cutlery in the utensil holder. All the cutlery should be oriented straight and facing forward''
    \item ``Fry two eggs at the front of the pan''
    \item ``Fry the spam on the pan and serve it on the plate''
    \item ``Weigh the shortest object and put it in the bin''
    \item ``Put the banana near where the other fruit are initially''
    \item ``Place the red block so that it's aligned with the other two blocks''
\end{tightlist}
\end{footnotesize}

In each task, the robot received the camera image and natural language command, estimated the scene geometry using a strategy similar to that in~\cite{m0m}, generated constraints and plans via PDDLStream~\cite{garrett2020PDDLStream}, and executed plans open-loop on hardware. No changes were made to the system across tasks.
\ours{} successfully completed all tasks, despite substantial variation in object configurations and goals.
Many tasks required adapting grasps (e.g., side grasps over top-down), managing collisions and reachability, or backtracking to move obstacles.
These results highlight \ours{}'s ability to flexibly and robustly integrate vision, language, and planning without any task-specific tuning.

While our system was able to solve versions of each of these tasks, we observed several failures on similar tasks and variants. The primary failure mode of OWL-TAMP was the same as in the simulated experiments: generating incorrect continuous constraints (e.g. the VLM generated a constraint for an object to be in collision with another object). A less common but still prevalent failure mode was generating correct continuous constraints that could not be satisfied with our finite sampling budget (e.g. a particular placement region is unreachable because the real-world robot has a rather limited workspace).
The majority of failures of the overall system were caused by common errors made by third-party real-world perception and control modules such as false negatives when segmenting objects and minor collisions due to imperfect trajectory tracking.

%% file: limitations.tex
\begin{figure}[t]
\vspace{0.25cm} 
\centering
\includegraphics[width=\linewidth]{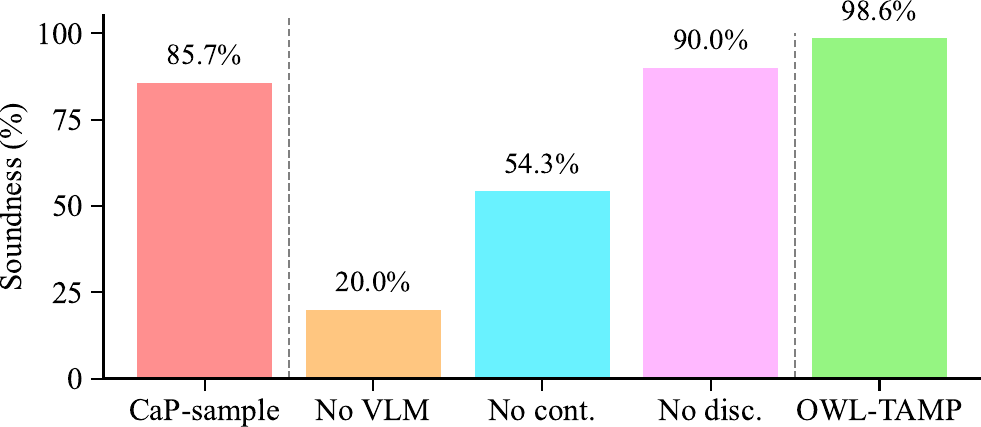}
\caption{
\textbf{Soundness rates on 7 tasks.}
Average soundness rate ($1 - \frac{\text{\# false positives}}{\text{\# total trials}}$) across tasks requiring non-trivial discrete or continuous constraint generation. A false positive occurs when success is declared despite the output plan violating task intent.
\(\uparrow\) indicates fewer false positives. Failed tasks (the method correctly declares failure) are counted as true positives.
%
}
\label{fig:soundness_results}
\vspace{-2em}
\end{figure}

\section{Conclusion}
\label{sec:conclusion}
We presented \ours, a system that uses VLMs to generate discrete and continuous constraints that enable a base TAMP system to interpret and accomplish open-world language instructions.
We demonstrated that \ours{} solves a wide range of complex, long-horizon language-based manipulation tasks more reliably than existing baselines, and that it applies across different robot embodiments in simulation and the real world.


\section{Limitations}
\label{sec:limitations}
Our system relies entirely on the VLM to generate constraints that are both syntactically and semantically correct.
While the TAMP system is capable of backtracking internally, our current implementation cannot recover from errors
in the generated constraints themselves.
Another limitation is our assumption of a set of primitive parameterized robot controllers (e.g., moving, rigid grasping, placement) and a library of primitive helper functions --- common in planning and foundation model-based code-generation works --- which \ours{} uses to generate plans and continuous constraints.
Despite these limitations, our results highlight the promise of combining large-scale pretrained models with structured planners, and we hope this work inspires further progress toward flexible, capable robot manipulation systems.


%% file: acknowledgements.tex
\section*{Acknowledgements}
Work for this project was partially conducted at NVIDIA Research.
We gratefully acknowledge support from NSF grant 2214177; from AFOSR grant FA9550-22-1-0249; from ONR MURI grants N00014-22-1-2740 and N00014-24-1-2603; from ARO grant W911NF-23-1-0034; from the MIT Quest for Intelligence; and from the Boston Dynamics Artificial Intelligence Institute.

%% file: appendix.tex
\appendix


\section{Appendix}
\label{appendix:additional-experiment-details}



We first provide details of how we implement OWL-TAMP (Appendix~\ref{appendix:method-details}), including details about its integration with standard TAMP systems, generation of discrete and continuous constraints.
Subsequent sections provide more specific details around prompting the VLM (Appendix~\ref{subsec:appendix-continuous-constraint-helpers},~\ref{subsec:appendix-owl-tamp-prompting-details}, and~\ref{appendix:cap-prompt}).
For details beyond those presented here, we refer the reader to our open-source code release, which we will make available after acceptance.

\subsection{Additional Method Implementation Details}
\label{appendix:method-details}
We provide a detailed explanation of how we implement the search-then-sample TAMP system used in our simulation experiments (Section~\ref{subsec:sim-results}).

\subsubsection{Search-then-Sample TAMP System Details} 
\label{subsec:appendix-tamp-details}
In our simulation experiments, we implement a relatively simple variant of the search-then-sample TAMP approach used in several recent works~\citep{srivastava2014combined,bilevel-planning-blog-post,silver2023predicate,kumar2023learning}.
We use this as the underlying TAMP systems for all methods that require one (i.e., all methods except `CaP' and `CaP-sample').
We implement versions of the same predicates and operators described in Section~\ref{sec:formulation} (including a version of \texttt{transport} that we call \texttt{Pour}), though we omit the \texttt{move} operator as movement is performed automatically as part of \texttt{attach} and \texttt{detach} (as illustrated by our prompts in Appendix~\ref{subsec:appendix-owl-tamp-prompting-details} below). 

We manually define the initial state of all tasks in terms of literals involving these predicates; the robot always begins each task at the same initial configuration, and with the \texttt{HandEmpty()} predicate set to \texttt{True}.
Additionally, we associate each parameterized action with a natural language description to make it easier for the VLM to perform discrete constraint generation (e.g., for the \texttt{detach} operator, the description is something like: ``places object \texttt{o} stably atop a surface; you can specify a description of how this placement should be performed by filling in the \texttt{description} parameter accordingly'').
Each operator is linked to one particular low-level controller included with the RAVENS-YCB environment~\cite{zeng2020transporter}.

We define an associated sampler for each of these parameterized actions. Each sampler takes in the current state of the task, as well as the action's discrete arguments, and defines a distribution over the action's continuous parameters.
These samplers are setup to define broad distributions (e.g.. the sampler for the \texttt{detach} operator simply tries to find a pose somewhere broadly above the surface to be detached onto, but roughly within the confines of the x and y boundaries of the surface's axis-aligned bounding box), though we do modify these slightly depending on the task (e.g. the sampler for the `Coffee' and `Mug' tasks defines particular distributions on the orientations the mug and/or cutlery should be \texttt{detach}-ed).

Given these components, we can adopt the following hierarchical search-then-sample planning strategy to achieve a symbolic predicate goal $G$ (derived from a plan sketch via the procedure listed in Appendix~\ref{subsec:appendix-discrete-constraint-generation} from initial state $s_0$\footnote{see~\citep{bilevel-planning-blog-post} or \citep{silver2023predicate} for a fuller presentation of the search-then-sample TAMP strategy we employ.}:
\begin{enumerate}
    \item Compute a new task plan (sequence of ground operators) that achieves $G$ from $s_0$
    \item For $b$ within the backtracking budget:
    \begin{enumerate}
            \item For each ground operator in this plan:
            \begin{enumerate}
                \item If the preconditions do not hold, break.
                \item For $i$ within the sampling budget:
                \begin{enumerate}
                    \item Call the associated sampler to get continuous parameters.
                    \item Use these to instantiate the controller associated with the operator.
                    \item Execute the controller within the planning model (i.e., the simulator)
                    \item Check that the resulting state satisfies all the constraints, including manually-engineered and VLM-generated constraints.
                \end{enumerate}
            \end{enumerate}
    \end{enumerate}
\end{enumerate}

We perform initial task planning via a simple $A^*$ search. During backtracking (i.e., when the sampling budget is exhausted for the first time and a new task plan is required), 
we employ a set of manually-engineered strategies to modify the task plan based on the most-recent failed operator (e.g. if the most recent-failed operator is a \texttt{detach} that was attempting to place an object atop a particular surface, and there are other objects atop that surface already, we randomly append a \texttt{attach} \texttt{detach} sequence to move one of those objects to a different part of the table).


\subsubsection{Discrete Constraint Generation Details}
\label{subsec:appendix-discrete-constraint-generation}

Given a goal described in natural language $L$ (Section~\ref{sec:formulation}), we first prompt a VLM to generate a \textit{plan sketch} (i.e., a partial plan) that serves as a discrete constraint on the space of TAMP solutions (Section~\ref{subsec:action-ordering}).
To enable this, we associate a natural language description of each available action with that particular action.
Although we could directly prompt a VLM for relevant actions and goals, without a list of candidates, the VLM is likely to be syntactically and semantically inaccurate.
Instead, we propose to first ground the set of {\em reachable} actions $A$ and literals $\Lambda$ (i.e., grounded predicates) available to the TAMP system before prompting the VLM to return values in these sets.
We use {\em relaxed planning}~\cite{bonet1999planning,bonet2001planning} from the initial state $s_0$ to simultaneously ground~\cite{HoffmannN01,edelkamp2000exhibiting,helmert2006fast} and explore the sets of reachable actions $A$ and literals $\Lambda$.
When instantiating continuous parameters, we use placeholder values, such as {\em optimistic} values~\cite{garrett2018stripstream,garrett2018sampling,garrett2020PDDLStream,garrett2021sampling}, to ensure a finite set of actions are instantiated.
Similarly, we use placeholders for \pddl{description} parameters.

\begin{algorithm}[h] 
\begin{small}
  \caption{VLM Task Reasoning} 
  \label{alg:planning-appendix}
  \begin{algorithmic}[1] 
    \Procedure{vlm-task-reasoning}{$s_0, {\cal A}, {\cal I}, L$}
        \State $A \gets \proc{ground-actions}(s_0, {\cal A})$
        \State $\Lambda \gets s_0 \cup \{l \mid a \in A.\; l \in e.\id{eff}\}$
        \State $[a_1, ..., a_n, l_1, .., l_m] \gets \proc{query-vlm}(\text{``What partial plan} \newline \text{using actions } \{A\} \text{ for goal literals } \{\Lambda\} \text{ achieves goal } \{L\}\text{?''})$
        \For{$i \in [1, n-1]$}
            \State $a_{i}.\id{eff} \gets a_{i}.\id{eff} \cup \{\pddl{Executed}(i)\}$
            \State $a_{i+1}.\id{pre} \gets a_{i+1}.\id{pre} \cup \{\pddl{Executed}(i)\}$
        \EndFor
        \State $a_{n}.\id{eff} \gets a_{n}.\id{eff} \cup \{\pddl{Executed}(n)\}$
        \State $G \gets \{l_1, .., l_m\}$
        \State \Return $\proc{solve-tamp}(s_0, A, G \cup \{\pddl{Executed}(n)\})$ 
    \EndProcedure
\end{algorithmic}
\end{small}
\end{algorithm}

Algorithm~\ref{alg:planning-appendix} presents the VLM partial plan generation pseudocode.
It takes in a TAMP problem $\langle s_0, {\cal A}, \mathcal{I}, L \rangle$, where $L$ is a text goal description and \(\mathcal{I}\) is a set of initial image observations of the scene (see Section~\ref{sec:formulation}).
It first grounds the set of actions $A$ reachable from $s_0$ using \proc{ground-actions}.
Then, it accumulates the set of reachable literals $\Lambda$ by taking the effects of all actions $A$.
These sets can be filtered by action or predicate type if it is desired to focus VLM assistance on specific aspects of the planning problem.
Then, it prompts \proc{query-vlm} for a partial plan $[a_1, ..., a_n, l_1, ..., l_k]$ using actions $a_i \in A$ and goal literals $l_j \in \Lambda_m$ that achieve the goal description $g$.
Importantly, we have the VLM fill in the description parameter $d$ for each of these actions.
We then transform the original TAMP problem to force solutions to admit the partial plan as a subsequence.
Specifically, we create a predicate \proc{Executed} that models whether the $i$th action in the plan was executed and add $\proc{Executed}$ to the effects of action $a_i$ and the preconditions of action $a_{i+1}$.
Finally, we make the planning goal be $G = \{l_i, ..., l_m\} \subseteq L$ and $\proc{Executed}(n)$, 
which indicates that all actions have been executed and solve the transformed TAMP problem with a generic TAMP algorithm.

\subsubsection{\ours{} Constraint Generation Example Walkthrough}
\label{subsec:appendix-owl-tamp-walkthru}


We walk through a concrete example of generating both discrete and continuous constraints for the `Mug1' task (Figure~\ref{fig:ravens-tasks}a), where the natural language goal is \(L\) = ``Setup the mug so it's upright, then put whatever object that fits inside of it".

First, we prompt the VLM to generate discrete constraints given the task goal. It generates the following partial plan, along with language descriptions filled in for each action:
\begin{lstlisting}[style=pythonstyle]
('Pick(mug)', 'grasp the mug securely to lift it from the table.')
('Place_Ontop(mug, table)', 'place the mug upright on the table to ensure it is stable.')
('Pick(fork)', 'grasp the fork securely to lift it from the table.')
('Place_Inside(fork, mug)', 'carefully place the fork inside the upright mug.')
('achieve_goal(mug, fork)', 'the goal is achieved when the mug is upright and the fork is inside it.')
\end{lstlisting}

Next, we prompt the VLM to provide continuous constraints for the \texttt{achieve\_goal} operator at the end of the plan. This corresponds to providing continuous constraints that correspond to the task's goal description.
\begin{lstlisting}[style=pythonstyle]
# Goal Check Functions:
def goal_check0() -> bool:
    upright_mug = abs(mug.pose.roll) < 0.1 and abs(mug.pose.pitch) < 0.1 and abs(mug.pose.yaw) < 0.1
    return upright_mug

def goal_check1() -> bool:
    inside_mug_bounds = modify_pose_bounds_to_be_ inside_object(init_state, env, init_bounds, mug.category)
    return position_within_bounds(fork.pose, inside_mug_bounds)
\end{lstlisting}

We then prompt the VLM to generate continuous constraints for each operator given these generated constraint functions as input.
We start from the beginning of the plan.
The \texttt{Pick(mug)} action (which corresponds to \texttt{attach}) does not have any effects or constraint predicates that depend on the VLM, so we skip this action.
The \texttt{Place\_Ontop(mug, table)} action ((which corresponds to \texttt{detach}) does feature a \texttt{VLMPose} constraint. So we prompt the VLM to generate continuous constraints that implement it based on the language description as well as the goal generated constraints.
\begin{lstlisting}[style=pythonstyle]
# Goal Check Functions:
def goal_check2() -> bool:
    upright_mug = abs(mug.pose.roll) < 0.1 and abs(mug.pose.pitch) < 0.1 and abs(mug.pose.yaw) < 0.1
    on_table_bounds = modify_pose_bounds_to_be_ontop _of_object(init_state, env, init_bounds, mug.category, 'table')
    return upright_mug and position_within_bounds(mug.pose, on_table_bounds)
\end{lstlisting}

The \texttt{Place\_Inside(fork, mug)} action (which also corresponds to \texttt{detach}) does not have any effects or constraint predicates that depend on the VLM, so we skip this action as well.
Finally, the \texttt{Place\_Inside(fork, mug)} action does have a \texttt{VLMPose} constraint, so we once again prompt the VLM, and obtain the following continuous constraints for this action:

\begin{lstlisting}[style=pythonstyle]
# Goal Check Functions:
def goal_check2() -> bool:
    inside_mug_bounds = modify_pose_bounds_ to_be_inside_object(init_state, env, init_bounds, mug.category)
    return position_within_bounds(fork.pose, inside_mug_bounds)
\end{lstlisting}

Note that in this particular case, the VLM only checks that the fork is inside the mug, which is a constraint that is built into the TAMP system itself (via the \texttt{Inside} predicate that is an effect of the \texttt{Place\_Inside} action).

Now, we have a partial plan along with continuous constraints for particular actions. We call our underlying TAMP system to attempt to satisfy these, as well as its own internal constraints (Appendix~\ref{subsec:appendix-tamp-details}).

\subsection{Additional RAVENS-YCB Manipulation Experiment Details}
\label{appendix:ravens-details}

\begin{figure*}[t]
    \centering
    \includegraphics[width=\textwidth]{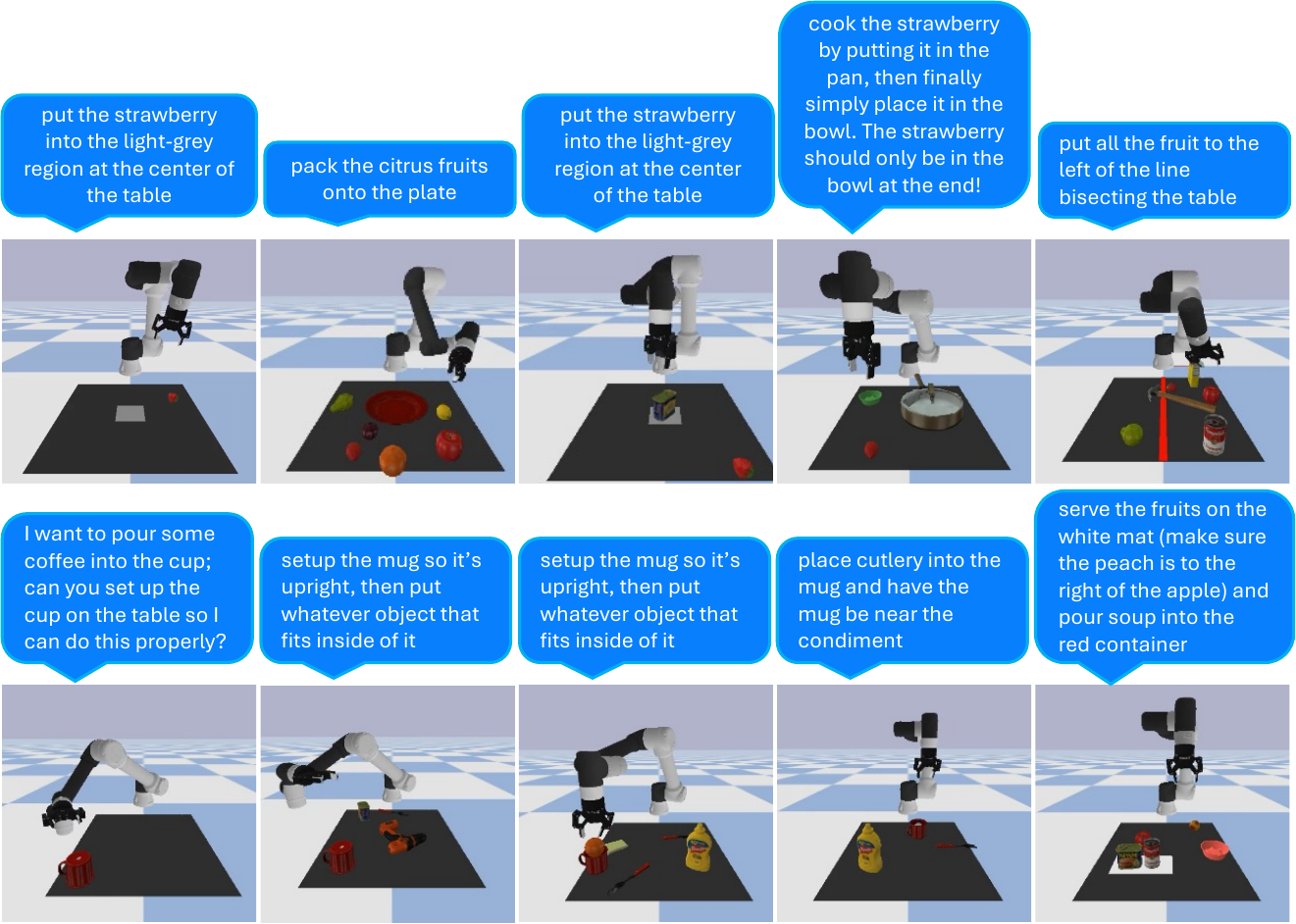}
    \caption{
    \small{\textbf{Ravens YCB-Manipulation tasks visualization.} From top left to bottom right: visualizations of the goal string and an example initial state for the `Berry1', `Citrus', `Berry2', `BerryCook', `FruitSort', `Coffee', `Mug1', `Mug2', `Mug3', and `SoupPour' tasks.}
    }
    \label{fig:all-sim-tasks}
\end{figure*}

We provide additional details on the tasks we ran experiments on in the simulated RAVENS-YCB Manipulation environment (Section~\ref{sec:experiments}) and discuss the Code as Policies baseline implementation in Appendix~\ref{subsec:cap-details}.

A visualization of an initial state in each of our tasks is shown in Figure~\ref{fig:all-sim-tasks}. In each of these tasks, the robot had access to three parameterized actions: \texttt{Pick($o, g$)} where $o$ is an object and $g$ is a 6-DOF grasp pose consisting of $[x, y, z, \id{roll}, \id{pitch}, \id{yaw}]$ in the world coordinate frame, \texttt{Place$(o, s, p)$} where $o$ is the object to be placed, $s$ is the surface or container to place atop or inside, and $p$ is the 6D placement pose at which the robot will move and simply open its gripper, and \texttt{Pour$(o, s, t)$}, where $o$ is the object to be poured from, $s$ is the surface or container to pour onto or into, and $t$ is a 4D vector consisting of a 3D position and a pitch angle at which to tip the hand to accomplish the pour.
For each random seed, the positions of all objects except the `table\_surface' object in the scene are randomized unless otherwise stated.

\begin{tightlist}
    \item \textit{Berry1}: Goal description: ``put the strawberry onto the light-grey region at the center of the table''. Objects: \texttt{strawberry, light\_grey\_region, table\_surface}. The position of the \texttt{light\_grey\_region} is held constant at the center of the table. The optimal solution to this task requires a sequence of 2 actions.
    \item \textit{Citrus}: Goal description: ``pack the citrus fruit onto the plate''. Objects: \texttt{strawberry, apple, pear, lemon, orange, plum, table\_surface}. The \texttt{plate} object is always set to be at the center of the table in the initial state. The optimal solution to this task requires a sequence of 4 actions.
    \item \textit{Berry2}: Goal description: ``put the strawberry onto the light-grey region at the center of the table''. Objects: \texttt{strawberry, light\_grey\_region, potted\_meat\_can, table\_surface}. The only object whose position is randomized in the initial state is the \texttt{strawberry}: the \texttt{potted\_meat\_can} is always set to totally obstruct the surface of the \texttt{light\_grey\_region} in the center of the table: it must be moved out of the way to make room to place the \texttt{strawberry} and successfully accomplish the goal description. The optimal solution to this task requires a sequence of 4 actions.
    \item \textit{BerryCook}: Goal description: ``Cook the strawberry by putting it in the pan, then finally simply place it in the bowl. The strawberry should only be in the bowl at the end!''. Objects: \texttt{strawberry, skillet, bowl, table\_surface}. The only object whose position is randomized in the initial state is the \texttt{strawberry}. The optimal solution to this task requires a sequence of 4 actions.
    \item \textit{FruitSort}: Goal description: ``Put all the fruit to the left of the line bisecting the table''.  Objects: \texttt{pear, sugar\_box, hammer, tomato\_soup\_can, strawberry, apple, red\_line, table\_surface}. The position of the \texttt{red\_line} object is kept constant at the center of the table. The optimal solution to this task requires a sequence of 6 actions.
    \item \textit{Coffee}: Goal description: ``I want to pour some coffee into the cup; can you set up the cup on the table so I can do this properly?'' (requires reorienting the cup so that it is placed `right-side-up' on the table). Objects: \texttt{mug, table\_surface}. The optimal solution to this task requires a sequence of 2 actions.
    \item \textit{Mug1}: Goal description: ``Setup the mug so it's upright, then put whatever object that fits inside of it''. Objects: \texttt{fork, power\_drill, potted\_meat\_can, mug, table\_surface}. The initial position of the \texttt{mug} is fixed. The optimal solution to this task requires a sequence of 4 actions.
    \item \textit{Mug2}: Goal description: ``Place cutlery inside the mug and then place the mug itself on the table near the condiment'' (the mug's opening is obstructed by a large orange, which must be moved out of the way). Objects: \texttt{fork, knife, sponge, strawberry, orange, mustard\_bottle, mug, table\_surface}. The initial position of the \texttt{mug} and \texttt{orange} are fixed such that the \texttt{orange} is always obstructing the mouth of the \texttt{mug}. The optimal solution to this task requires a sequence of 8 actions.
    \item \textit{Mug3}: Goal description: ``place cutlery into the mug and have the mug be near the condiment''. This is the same as \textit{CutleryInMug2}, except in the initial state the mug is not obstructed by an orange, but rather has a ball stuck inside it. This must be removed by `pouring' it out of the cup itself). 
    Objects: \texttt{fork, golf\_ball, mustard\_bottle, mug, table\_surface}. The initial position of the \texttt{mug} and \texttt{golf\_ball} are fixed such that the \texttt{golf\_ball} is always inside the \texttt{mug}.
    The optimal solution to this task requires a sequence of 8 actions.
    \item \textit{SoupPour}: Goal description: ``Serve the fruits on the white mat (make sure the peach is to the right of the apple'' and pour soup into the red container''. The white mat is originally obstructed by the soup can as well as a spam can, which must be moved out of the way to successfully place the fruits. Objects: \texttt{white\_mat, tomato\_soup\_can, potted\_meat\_can, bowl, apple, peach, table\_surface}. The only object positions that are randomized in the initial state are that of the \texttt{apple} and the \texttt{peach}. The optimal solution to this task requires a sequence of 10 actions.
\end{tightlist}

\subsubsection{Code as Policies Implementation Details}
\label{subsec:cap-details}
Following recent work~\citep{curtis2024trustproc3ssolvinglonghorizon}, we adapt Code as Policies to our RAVENS YCB-Manipulation domain by having it output a Language Model Program (LMP) function that --- given an initial state --- directly outputs a full plan with discrete actions and continuous parameters. We provide it with all the same helper functions used for continuous parameter generation that our method has access to. 
We also provide it with few-shot example solutions to the same 3 example problems as we provide in the continuous constraint generation prompt of our approach.
Additionally, we provide it with one sampler per controller that it can use to generate continuous parameters for each respective controller.
We provide the full detailed prompt we use for this method below in Appendix~\ref{appendix:cap-prompt}.

\begin{table*}[t]
\centering
\resizebox{\textwidth}{!}{%
\begin{tabular}{lcccccccccc}
    \toprule
    \multicolumn{1}{c}{} & \multicolumn{10}{c}{\textbf{Tasks}}\\ 
    \cmidrule{2-11}
    \textbf{Method} & \textbf{Berry1} & \textbf{Citrus} & \textbf{Berry2} & \textbf{BerryCook} & \textbf{FruitSort} & \textbf{Coffee} & \textbf{Mug1} & \textbf{Mug2} & \textbf{Mug3} & \textbf{SoupPour} \\
    \midrule
    CaP-sample & \makecell{83.30\\ $\pm$110.55} & \makecell{2185.30\\ $\pm$546.48} & \makecell{2353.50\\ $\pm$363.06} & \makecell{2606.00\\ $\pm$25.00} & \makecell{0.00\\ $\pm$0.00} & \makecell{255.60\\ $\pm$161.55} & \makecell{2519.90\\ $\pm$3.45} & \makecell{500.90\\ $\pm$755.41} & \makecell{2557.10\\ $\pm$9.09} & \makecell{1770.70\\ $\pm$874.10} \\
    \midrule
    No VLM & \makecell{34.70\\ $\pm$19.45} & \makecell{247.70\\ $\pm$81.38} & \makecell{619.40\\ $\pm$28.15} & \makecell{15.90\\ $\pm$13.46} & \makecell{0.00\\ $\pm$0.00} & \makecell{0.00\\ $\pm$0.00} & \makecell{856.40\\ $\pm$661.17} & \makecell{819.50\\ $\pm$374.59} & \makecell{320.20\\ $\pm$186.27} & \makecell{113.50\\ $\pm$57.43} \\
    \addlinespace[0.3em]
    No disc. & \makecell{42.10\\ $\pm$16.74} & \makecell{303.70\\ $\pm$85.50} & \makecell{729.30\\ $\pm$126.84} & \makecell{47.00\\ $\pm$28.48} & \makecell{0.00\\ $\pm$0.00} & \makecell{0.00\\ $\pm$0.00} & \makecell{1636.90\\ $\pm$349.74} & \makecell{1293.00\\ $\pm$208.14} & \makecell{915.30\\ $\pm$244.54} & \makecell{208.70\\ $\pm$139.25} \\
    \addlinespace[0.3em]
    No cont. & \makecell{35.10\\ $\pm$19.37} & \makecell{349.40\\ $\pm$244.23} & \makecell{193.50\\ $\pm$115.46} & \makecell{41.00\\ $\pm$25.54} & \makecell{308.30\\ $\pm$167.53} & \makecell{132.40\\ $\pm$78.32} & \makecell{376.80\\ $\pm$299.85} & \makecell{2500.00\\ $\pm$0.00} & \makecell{1478.70\\ $\pm$512.81} & \makecell{1170.60\\ $\pm$362.64} \\
    \addlinespace[0.3em]
    No back. & \makecell{37.70\\ $\pm$16.59} & \makecell{229.10\\ $\pm$105.03} & \makecell{181.20\\ $\pm$109.62} & \makecell{36.40\\ $\pm$17.54} & \makecell{348.20\\ $\pm$86.87} & \makecell{167.60\\ $\pm$77.11} & \makecell{348.30\\ $\pm$145.08} & \makecell{369.50\\ $\pm$150.42} & \makecell{500.00\\ $\pm$0.00} & \makecell{450.00\\ $\pm$113.11} \\
    \midrule
    OWL-TAMP & \makecell{37.70\\ $\pm$16.59} & \makecell{250.30\\ $\pm$140.93} & \makecell{181.20\\ $\pm$109.62} & \makecell{101.90\\ $\pm$67.60} & \makecell{372.40\\ $\pm$114.46} & \makecell{167.60\\ $\pm$77.11} & \makecell{696.30\\ $\pm$508.87} & \makecell{710.00\\ $\pm$461.70} & \makecell{1296.30\\ $\pm$607.64} & \makecell{1335.10\\ $\pm$409.67} \\
    \midrule\midrule
    Manual & \makecell{42.10\\ $\pm$16.74} & \makecell{303.70\\ $\pm$85.50} & \makecell{120.60\\ $\pm$50.45} & \makecell{43.30\\ $\pm$12.77} & \makecell{271.70\\ $\pm$99.02} & \makecell{197.80\\ $\pm$125.70} & \makecell{365.10\\ $\pm$312.24} & \makecell{657.70\\ $\pm$666.77} & \makecell{264.40\\ $\pm$202.57} & \makecell{832.80\\ $\pm$785.73} \\
    \bottomrule
\end{tabular}}
\caption{\textbf{Number of samples on all tasks}. We report the average number of continuous parameter sampling attempts for each task across $10$ random seeds; the number of samples within the 95\% confidence interval is reported after the $\pm$ sign for every entry. We also include a row (`Manual') reporting results on a variant of~\ours{} where we manually specify ground-truth discrete and continuous constraints to serve as an approximate lower bound on the number of samples our approach could achieve.}\label{tab:method_samples}
\end{table*}

\subsection{Additional Real-World Experiment Details}
We attempted the following goal descriptions on the real robot:
\begin{footnotesize}
\begin{tightlist}
    \item ``Put the orange and apple on the plate.''
    \item ``Place the strawberry and lime each in the bin that matches their color.''
    \item ``Stack the blocks into a tower by increasing hue.''
    \item ``Put the apple left of the plate and the orange on the table surface behind of the plate.''
    \item ``Put the orange on the far right of the table and the apple on the far left''.
    \item ``Put the orange where the apple is initially''.
    \item ``Clean the plate'' (a sponge is among several other objects present on a tabletop, and the robot must put the sponge atop the plate)
    \item ``Throw away anything not vegan in the purple bin'' (objects on the table include a milk carton, apple, spam can, and water bottle).
    \item ``Put the green block between the blue and red ones'' 
    \item ``Put the blue block onto the plate'' (the plate is packed with distractors and the robot must make a tightly-constrained placement).
    \item ``Setup the cutlery for someone to eat a meal from the plate. All the cutlery should be close to and lined-up with the plate, and should be oriented so each is straight and facing forwards, though you should pick which side of the plate each of the items are on'' (there are two pieces of fruit, and two similarly-colored blocks that must be disambiguated).
    \item ``Fit one of the fruit in the cup'' (only one of 4 available fruits is small enough to fit in the provided mug)
    \item ``Put the brownie ingredients in front of the pan'' (only 2 of the available items are related to brownies)
    \item ``Place the cutlery in the utensil holder. All the cutlery should be oriented straight and facing forward''
    \item ``Fry two eggs at the front of the pan''
    \item ``Fry the spam on the pan and serve it on the plate''
    \item ``Weigh the shortest object and put it in the bin''
    \item ``Put the banana near the other fruit''
    \item ``Place the red block so that it’s aligned with the other two blocks''
\end{tightlist}
\end{footnotesize}

Please see our supplementary video to see the real robot execution.

\subsection{Additional Experimental Results}
\label{appendix:additional-results}

\begin{table*}[t!]
\centering
\resizebox{\textwidth}{!}{%
\setlength{\tabcolsep}{4pt}
\begin{tabular}{lcccccccccc}
    \toprule
    \multicolumn{1}{c}{} & \multicolumn{10}{c}{\textbf{Tasks}}\\ 
    \cmidrule{2-11}
    \textbf{Method} & \textbf{Berry1} & \textbf{Citrus} & \textbf{Berry2} & \textbf{BerryCook} & \textbf{FruitSort} & \textbf{Coffee} & \textbf{Mug1} & \textbf{Mug2} & \textbf{Mug3} & \textbf{SoupPour} \\
    \midrule
    CaP & \makecell{10.03\\ $\pm$0.32\\ (47.38\%)} & \makecell{21.27\\ $\pm$1.10\\ (25.02\%)} & \makecell{12.63\\ $\pm$1.13\\ (42.76\%)} & \makecell{13.77\\ $\pm$0.95\\ (45.88\%)} & \makecell{19.50\\ $\pm$0.67\\ (26.14\%)} & \makecell{9.66\\ $\pm$0.80\\ (47.61\%)} & \makecell{19.77\\ $\pm$2.84\\ (46.16\%)} & \makecell{23.25\\ $\pm$1.65\\ (30.42\%)} & \makecell{16.37\\ $\pm$1.49\\ (38.14\%)} & \makecell{19.33\\ $\pm$1.17\\ (42.88\%)} \\
    \addlinespace[0.3em]
    CaP-sample & \makecell{22.40\\ $\pm$12.70\\ (21.22\%)} & \makecell{248.49\\ $\pm$54.38\\ (2.14\%)} & \makecell{245.31\\ $\pm$35.87\\ (2.20\%)} & \makecell{262.43\\ $\pm$14.12\\ (2.41\%)} & \makecell{20.88\\ $\pm$2.05\\ (24.41\%)} & \makecell{40.39\\ $\pm$16.21\\ (11.39\%)} & \makecell{281.74\\ $\pm$2.99\\ (3.24\%)} & \makecell{75.61\\ $\pm$78.73\\ (9.35\%)} & \makecell{274.50\\ $\pm$11.97\\ (2.27\%)} & \makecell{187.90\\ $\pm$83.47\\ (4.41\%)} \\
    \midrule
    No VLM & \makecell{11.90\\ $\pm$1.35\\ (8.85\%)} & \makecell{41.59\\ $\pm$4.85\\ (4.53\%)} & \makecell{47.66\\ $\pm$3.20\\ (1.60\%)} & \makecell{15.96\\ $\pm$1.23\\ (29.77\%)} & \makecell{16.56\\ $\pm$0.14\\ (24.60\%)} & \makecell{4.92\\ $\pm$0.04\\ (10.25\%)} & \makecell{59.43\\ $\pm$31.91\\ (4.11\%)} & \makecell{78.72\\ $\pm$18.85\\ (6.36\%)} & \makecell{36.03\\ $\pm$11.85\\ (8.04\%)} & \makecell{21.28\\ $\pm$2.54\\ (20.83\%)} \\
    \addlinespace[0.3em]
    No sample & \makecell{44.00\\ $\pm$4.03\\ (85.97\%)} & \makecell{75.17\\ $\pm$5.26\\ (72.29\%)} & \makecell{59.47\\ $\pm$4.69\\ (83.77\%)} & \makecell{54.93\\ $\pm$4.68\\ (82.13\%)} & \makecell{81.19\\ $\pm$10.40\\ (74.17\%)} & \makecell{39.95\\ $\pm$4.28\\ (85.49\%)} & \makecell{55.07\\ $\pm$3.43\\ (73.59\%)} & \makecell{80.00\\ $\pm$9.63\\ (69.38\%)} & \makecell{46.22\\ $\pm$2.57\\ (70.58\%)} & \makecell{91.46\\ $\pm$29.02\\ (82.12\%)} \\
    \addlinespace[0.3em]
    No disc. & \makecell{44.92\\ $\pm$4.07\\ (70.03\%)} & \makecell{91.62\\ $\pm$6.22\\ (50.92\%)} & \makecell{80.87\\ $\pm$6.83\\ (33.37\%)} & \makecell{44.50\\ $\pm$4.38\\ (69.50\%)} & \makecell{36.10\\ $\pm$2.01\\ (59.87\%)} & \makecell{18.92\\ $\pm$2.96\\ (72.29\%)} & \makecell{118.42\\ $\pm$16.71\\ (20.71\%)} & \makecell{120.94\\ $\pm$9.99\\ (29.63\%)} & \makecell{81.85\\ $\pm$10.58\\ (26.85\%)} & \makecell{96.95\\ $\pm$6.37\\ (72.54\%)} \\
    \addlinespace[0.3em]
    No cont. & \makecell{18.15\\ $\pm$1.84\\ (40.92\%)} & \makecell{54.53\\ $\pm$13.09\\ (16.21\%)} & \makecell{36.91\\ $\pm$5.61\\ (29.48\%)} & \makecell{24.30\\ $\pm$1.91\\ (29.40\%)} & \makecell{57.03\\ $\pm$9.88\\ (15.68\%)} & \makecell{25.73\\ $\pm$4.13\\ (29.93\%)} & \makecell{52.26\\ $\pm$16.08\\ (14.61\%)} & \makecell{136.88\\ $\pm$5.72\\ (5.72\%)} & \makecell{102.53\\ $\pm$22.20\\ (6.04\%)} & \makecell{91.00\\ $\pm$20.21\\ (9.42\%)} \\
    \addlinespace[0.3em]
    No back. & \makecell{48.69\\ $\pm$3.40\\ (71.92\%)} & \makecell{106.36\\ $\pm$24.85\\ (60.35\%)} & \makecell{101.84\\ $\pm$34.28\\ (72.59\%)} & \makecell{67.94\\ $\pm$15.09\\ (75.11\%)} & \makecell{149.53\\ $\pm$39.78\\ (66.52\%)} & \makecell{57.68\\ $\pm$8.20\\ (65.30\%)} & \makecell{87.31\\ $\pm$21.78\\ (58.23\%)} & \makecell{106.82\\ $\pm$6.13\\ (54.49\%)} & \makecell{69.59\\ $\pm$3.02\\ (52.24\%)} & \makecell{111.50\\ $\pm$32.13\\ (67.36\%)} \\
    \midrule
    OWL-TAMP & \makecell{46.92\\ $\pm$3.37\\ (74.64\%)} & \makecell{108.50\\ $\pm$24.61\\ (59.15\%)} & \makecell{101.88\\ $\pm$34.47\\ (72.56\%)} & \makecell{71.38\\ $\pm$15.31\\ (71.49\%)} & \makecell{154.32\\ $\pm$43.43\\ (64.46\%)} & \makecell{57.96\\ $\pm$7.69\\ (64.98\%)} & \makecell{113.18\\ $\pm$28.50\\ (44.92\%)} & \makecell{140.51\\ $\pm$23.06\\ (41.43\%)} & \makecell{118.38\\ $\pm$23.74\\ (30.71\%)} & \makecell{173.81\\ $\pm$49.00\\ (43.21\%)} \\
    \midrule\midrule
    Manual & \makecell{13.53\\ $\pm$2.36\\ (0.00\%)} & \makecell{44.42\\ $\pm$5.34\\ (0.00\%)} & \makecell{24.84\\ $\pm$2.78\\ (0.00\%)} & \makecell{19.78\\ $\pm$1.34\\ (0.00\%)} & \makecell{49.72\\ $\pm$3.59\\ (0.00\%)} & \makecell{22.15\\ $\pm$6.36\\ (0.00\%)} & \makecell{45.18\\ $\pm$16.74\\ (0.00\%)} & \makecell{79.91\\ $\pm$33.04\\ (0.00\%)} & \makecell{37.64\\ $\pm$10.40\\ (0.00\%)} & \makecell{81.38\\ $\pm$41.85\\ (0.00\%)} \\
    \bottomrule
\end{tabular}}
\caption{\textbf{Wall clock time on all tasks}. We report the average wall clock time in seconds each approach took to solve each task across $10$ random seeds; the time within the 95\% confidence interval is reported after the $\pm$ sign, and the average percentage of that time that was spent querying a foundation model is reported within parentheses '()' for every entry. We also include a row (`Manual') reporting results on a variant of~\ours{} where we manually specify ground-truth discrete and continuous constraints (and thus no need to query a foundation model for these) to serve as an approximate lower bound on the time our approach could achieve.}\label{tab:method_times}
\end{table*}


In this section, we explore experimental results related to how much computation and time our various methods took on each of the simulated tasks in the RAVENS YCB-Manipulation domain from Section~\ref{sec:experiments}. We also present a fine-grained table of soundness rate results used to construct Figure~\ref{fig:soundness_results} in Table~\ref{tab:soundness_results}.

Table~\ref{tab:method_samples} shows the number of samples required by the various methods to solve each of the RAVENS-YCB Manipulation tasks.
Note importantly that we do not adjust for success rates (depicted in Table~\ref{tab:method_env_results}) here: many approaches that achieve a low number of samples (e.g. CaP-sample or No VLM on FruitSort) here actually do so only because they in fact fail to translate that particular task into a form that they can attempt to solve via sampling.
\ours{} generally does not require significantly more samples than the `Manual' baseline (which serves as an approximate lower bound on the samples).
In two of the tasks where it does require substantially more samples than `Manual' (Mug3, SoupPour), \ours{} provides discrete constraints that yield an initial plan skeleton that is incorrect, necessitating backtracking within the TAMP system, which significantly increases the number of samples.

Table~\ref{tab:method_times} shows the wall clock time required by the various methods to solve each of the RAVENS-YCB Manipulation tasks.
Note once again that we do not adjust for success rates here.
In general, we see that our approach takes significantly more wall clock time than the `Manual' oracle, but we also see that a substantial portion of the time taken (around 50\% on average) is due to querying a foundation model.
We also see that the percentage of time spent querying increases roughly with the task complexity (left to right), which is what we would expect, since in the more complex tasks with more constraints, it is more challenging to find a satisfying sample.
These findings indicate that the overall runtime of our method could be significantly reduced by reducing the time taken for foundation model querying.

\begin{table}[t]
\centering
\begin{small}
\setlength{\tabcolsep}{4pt}
\begin{tabular}{lrrrrrrr}
\toprule
\multicolumn{1}{c}{} & \multicolumn{7}{c}{\textbf{Tasks}} \\ 
\cmidrule{2-8}
\textbf{Method} & {BerryCook} & {FruitSort} & {Coffee} & {Mug1} & {Mug2} & {Mug3} & {SoupPour} \\
\midrule
CaP-sample & 100\% & 100\% & 0\% & 100\% & 100\% & 100\% & 100\% \\
\midrule
No VLM & 0\% & 0\% & 0\% & 20\% & 0\% & 20\% & 100\% \\
No disc. & 40\% & 100\% & 90\% & 100\% & 100\% & 100\% & 100\% \\
No cont. & 100\% & 10\% & 60\% & 70\% & 100\% & 20\% & 20\% \\
\midrule
OWL-TAMP & 100\% & 100\% & 100\% & 100\% & 100\% & 100\% & 90\% \\
\bottomrule
\end{tabular}%
\end{small}
\vspace{5px}
\caption{\textbf{Soundness rates on select methods and tasks}. For tasks that require non-trivial discrete or continuous constraint generation, we present the `soundness rate' (i.e., 1 - $\frac{\text{\# false positives}}{\text{\# total trials}}$), where higher rates indicate that the method has fewer false positives.}
\label{tab:soundness_results}
\end{table}

\subsection{Real Robot System Implementation}
\label{appendix:robot-implementation-details}

For the real-world demos, we adapted the strategy of \citet{m0m}, which deploys TAMP without a priori object models by estimating collision, grasp, and placement affordances online.
Here, we used PDDLStream and its ``Adaptive'' solver~\cite{garrett2020PDDLStream} to implement the TAMP framework and algorithm within \ours{}, demonstrating \ours{}'s compatibility with multiple off-the-shelf TAMP systems. 
We deployed \ours{} in a replanning policy~\cite{garrett2020online}, where at each state, the robot observes the world with its head camera, segments and estimates the object geometries, plans a course of action, and executes its plan.
We used Grounding DINO~\cite{liu2023grounding} for object detection, segmentation, and association, RVT~\cite{goyal2023rvt} to infer grasp affordances, TRAC-IK~\cite{beeson2015trac} for inverse kinematics, and cuRobo~\cite{sundaralingam2023curobo} for motion planning.



\subsection{Helper Functions for Continuous Constraint Generation.}
\label{subsec:appendix-continuous-constraint-helpers}
We provide all methods (Section~\ref{sec:experiments}) access to the following helper functions to be used towards constraining the continuous parameters used to instantiate actions. 
We provide the name, parameters and docstring of each function exactly as below:

\begin{lstlisting}[style=pythonstyle]
def get_aabb_bounds
    """Given the state of a particular env, and an object_name that appears in
    this state, return tuples corresponding to the bounds of the axis-aligned
    bounding box of object_name in this state in the world frame.

    In particular, return the lower and upper bounds on the axis-aligned
    x, y, z values.
    """

def get_obj_center
    """Given the state of a particular env, return the pose of the object with
    object_name.

    The pose is a tuple of dim 6 corresponding to (x, y, z, roll, pitch,
    yaw).
    """

def modify_pose_bounds_to_be_behind_object
    """Given a tuple of initial bounds (init_bounds), return a modified set of
    bounds such that sampling randomly from the output bounds will ensure that
    a pose will be selected that is behind (on the table plane) the object with
    name `object_name`'s. For instance:
    modify_pose_bounds_to_be_behind _object(init_state, env, init_bounds,
    'hammer') will modify init_bounds such that they only contain poses that
    are behind the 'hammer' object on the table surface ahead the robot.

    Note that this does not constrain
    the pose's horizontal position (it may be anywhere on the table - in the
    left or right half - such that it's behind object_name).
    """

def modify_pose_bounds_to_be_in_front_of_object
    """Given a tuple of initial bounds (init_bounds), return a modified set of
    bounds such that sampling randomly from the output bounds will ensure that
    a pose will be selected that is to the in front of (on the table plane) the
    object with name `object_name`'s. For instance:
    modify_pose_bounds_to_be_in_front_of_ object(init_state, env, init_bounds,
    'hammer') will modify init_bounds such that they only contain poses that
    are in front of the 'hammer' object on the table surface ahead the robot.

    Note that this does not constrain
    the pose's horizontal position (it may be anywhere on the table - in the
    left or right half - such that it's in front of object_name).
    """

def modify_pose_bounds_to_be_left_of_object
    """Given a tuple of initial bounds (init_bounds), return a modified set of
    bounds such that sampling randomly from the output bounds will ensure that
    a pose will be selected that is to the left of (on the table plane) the
    object with name `object_name`'s. For instance:
    modify_pose_bounds_to_be_left _of_object(init_state, env, init_bounds,
    'hammer') will modify init_bounds such that they only contain poses that
    are to the left of the 'hammer' object on the table surface ahead the
    robot.

    Note that this does not constrain
    the pose's vertical position (it may be anywhere on the table - in the
    upper or lower half - such that it's to the left of object_name).
    """

def modify_pose_bounds_to_be_right_of_object
    """Given a tuple of initial bounds (init_bounds), return a modified set of
    bounds such that sampling randomly from the output bounds will ensure that
    a pose will be selected that is to the right of (on the table plane) the
    object with name `object_name`'s. For instance:
    modify_pose_bounds_to_be_right_ of_object(init_state, env, init_bounds,
    'hammer') will modify init_bounds such that they only contain poses that
    are to the right of the 'hammer' object on the table surface ahead the
    robot.

    Note that this does not constrain
    the pose's vertical position (it may be anywhere on the table - in the
    upper or lower half - such that it's to the right of object_name).
    """

def modify_pose_bounds_to_be_above_object
    """Given a tuple of initial bounds (init_bounds), return a modified set of
    bounds such that sampling randomly from the output bounds will ensure that
    a pose will be selected that is above (on the table plane) the object with
    name `object_name`'s. For instance:
    modify_pose_bounds_to_be_ above_object(init_state, env, init_bounds,
    'hammer') will modify init_bounds such that they only contain poses that
    are above the 'hammer' object on the table surface ahead the robot.

    Note that this does actually also constrain the pose's horizontal
    position and vertical positions so that it is directly above the
    object in question. Note also that this function might particularly
    useful to constraint pouring actions (because pouring must be done
    from above); though you will also likely have to apply an additional
    angular constraint (since this function doesn't apply any angular
    constraints on its own).
    """

def modify_pose_bounds_to_be_below_object
    """Given a tuple of initial bounds (init_bounds), return a modified set of
    bounds such that sampling randomly from the output bounds will ensure that
    a pose will be selected that is below (on the table plane) the object with
    name `object_name`'s. For instance:
    modify_pose_bounds_to_be below_object(init_state, env, init_bounds,
    'hammer') will modify init_bounds such that they only contain poses that
    are below the 'hammer' object on the table surface ahead the robot.

    Note that this does actually also constrain the pose's horizontal
    position and vertical positions so that it is directly below the
    object in question. Note also that this function might particularly
    useful to constraint pouring actions (because pouring must be done
    from above into a container that's below); though you will also
    likely have to apply an additional angular constraint (since this
    function doesn't apply any angular constraints on its own).
    """

def modify_pose_bounds_to_be_near_object
    """Given a tuple of initial bounds (init_bounds), return a modified set of
    bounds such that sampling randomly from the output bounds will ensure that
    a pose will be selected such that the distance of the pose from the object
    with name `object_name` will be within closeness_thresh along the x, y, and
    z axes respectively (note that the pose might have an L2) distance that's
    greater than that."""

def modify_pose_bounds_to_be_ontop_of_object
    """Assuming the init_bounds are on the pose (x, y, z, roll, pitch, yaw) of
    an object with name obj1_name, modify these such that the pose must be
    confined to be on top of the object with name obj2_name.

    Specifically, restrict the bounds to be within x and y of
    obj2_name's bounding-box, but have its z-position touching the top
    of the bounding box of obj2_name.

    IMPORTANT: use this only when trying to place an object atop another
    (e.g. atop a region, or a surface of another object). If you want to put
    something inside a container, use the
    modify_pose_bounds_to_be_ontop_of_object function instead.
    """

def modify_pose_bounds_to_be_inside_object
    """Assuming the init_bounds are on the pose (x, y, z, roll, pitch, yaw) of
    an object with name obj1_name, modify these such that the pose must be
    confined to be inside the object with name obj2_name.

    Specifically, restrict the bounds to be within x and y of
    obj2_name's bounding-box.

    IMPORTANT: use this only when trying to place an object inside a container
    (e.g. a cup, or vase, or 3D box). If you want to put something in a 2D
    region, use the modify_pose_bounds_to_be_ontop_of_object function instead.
    Also note that this function is generally not suitable to constrain
    pouring; it should generally be used when constraining placement!
    """

def position_within_bounds
    """Checks that the xyz position component of a 6-d pose is within specific
    bounds."""

def initialize_bounds_anywhere_on_object
    """Given obj, get its aabb and initialize bounds such that sampling within
    these bounds will yield a pose with a position atop obj and any arbitrary
    rotation."""

def sample_ravenpose_uniformly_within_bounds
    """Given obj, get its aabb and initialize bounds such that sampling within
    these bounds will yield a pose with a position atop obj and any arbitrary
    rotation."""

def modify_obj_pose
    """Modifies the pose of obj to new_pose."""
\end{lstlisting}

\subsection{\ours{} prompting details.} 
\label{subsec:appendix-owl-tamp-prompting-details}
As described in Section~\ref{sec:tamp-with-open-world-concepts}, our approach consists of an initial discrete constraint generation phase followed by continuous constraint generation based on a discrete partial plan skeleton with language parameters filled in (i.e., plan sketch).

We use separate prompts for both the discrete and continuous constraint generation. 
For both discrete and continuous constraint generation, we provide an image of the initial state of the task as part of the prompt.

Our discrete constraint generation prompt provides a single few-shot example of expected output on a particular task, and then requests the VLM to output a plan in a similar format for the current task.
It also leverages chain of thought prompting~\citep{wei2023chainofthoughtpromptingelicitsreasoning} to encourage the model to improve the output accuracy.
The full prompt we use is shown below: variables within curly brackets (\{\}) are filled-in dynamically depending on the task.

\begin{lstlisting}[style=promptstyle]
You are an expert-level robot task planning system whose job is to help a robot accomplish the following task: ''{task_str}''.

Here is the initial predicate state (i.e., the set of all ground atoms that are true) of this task. Note that an image corresponding to the environment in this state
is attached below:
{initial_preds}

Your job is to output a sequence of ground operators (i.e., a plan) that ideally achieve the goal from this initial state.
Your plan need not be perfect, but it should capture the critical objects and actions necessary to accomplish this task (e.g. 
if the task requires 4 objects being in a specific location, then you should take care to make sure the plan contains
an action to manipulate each of the 4 objects in turn).

Here are the unground operators with their descriptions.
{nsrts_description}

Here are all the ground operators available to you; each operator you use in your plan must be one of these.
{ground_operators}

Along with each operator in your plan, you should also output a natural language description of what that operator should 
do. This description can be as detailed as you like, and should explain any details relevant to completing the particular
ground operator successfully.

As an example, consider the example task ''serve the banana inside the blue thing''. Here, the bowl happens to be blue, and 
the initial state is:
OnTable(banana)
OnTable(bowl)
And the available ground operators are:
pick(banana)
pick(bowl)
pick(table)
place_ontop(banana, bowl)
place_inside(banana, bowl)
place_ontop(bowl, banana)
place_inside(bowl, banana)
place_ontop(banana, table)
place_inside(banana, table)
place_ontop(bowl, table)
place_inside(bowl, table)
place_ontop(table, bowl)
place_inside(table, bowl)

Given this, the output should be something like:
"""
In the initial state, there is a blue bowl on the table, and a banana atop the table. The banana is not in the bowl, and the task is to
move the banana into the bowl.
The main actions relevant to the task are `pick(banana)` and `place(banana, bowl)`. The goal involves a relationship between the banana and the bowl only.
All other objects can be ignored.
Plan:
pick(banana); make a stable grasp on the banana - try to make a top-down grasp for maximum likelihood of success
place(banana, bowl); place the banana stably so that it rests in the bowl - the banana is too large to fit inside the bowl if it is placed flatly: it needs to be reoriented to be upright so that it can fit into the bowl
achieve_goal(banana, bowl); the goal involves the banana being inside the bowl - this relationship is purely between the banana and bowl and doesn't involve/require any other objects.
"""
Notice how the plan ends in an `achieve_goal` operator. Every plan you output should end with such an operator, and the object arguments 
to this operator (i.e., `(banana, bowl)` in this case) should be all the objects necessary to decide whether or not the goal has been achieved 
(i.e., do your best not to include extraneous objects that are irrelevant to deciding whether the task goal has been achieved or not).

Please output your plan in the following format (do not include the angle brackets: those are just for illustrative purposes). Importantly, please do not list the plan with a numbered or bulleted list,
simply output each ground operator on a new line with no marking in front of the line as indicated below.
<description of the initial state and task in your own words>
<description of which objects and actions are particularly relevant to solving the task>
<description of any challenges or other important considerations/obstacles that might arise when solving the task>
Plan:
<ground_operator0>; <natural language description0>
<ground_operator1>; <natural language description1>
...
<ground_operatorm>; <natural language descriptionm>
\end{lstlisting}

Note that we ask the model to output a \texttt{achieve\_goal} operator at the end of the plan.
This is used by the continuous constraint generation procedure that follows: we generate continuous constraints for this action (the natural language description of the \texttt{achieve\_goal} action is effectively the task's goal description $g$) first --- which effectively corresponds to generating constraints for the task goal --- and then generate constraints for any previous operators in the plan by conditioning on these goal continuous constraints.

We prompt the VLM to generate goal continuous constraints by providing the helper functions listed above, available objects in the scene, as well as three few-shot examples (shown below) of outputs on three separate simple example problems:

\begin{lstlisting}[style=promptstyle]
To give you an idea of what your output function should look like, here is an example function generated for the task "put the lemon on the spoon and the banana on the table", where "lemon", "spoon", "banana", and "table" are all objects in that task/scene.

```python
def goal_check0() -> bool:
    ontop_spoon_bounds = modify_pose_bounds_to_ be_ontop_of_object(init_state, env, init_bounds, lemon.category, spoon.category)
    return position_within_bounds(lemon.pose, ontop_spoon_bounds)
```
```python
def goal_check1() -> bool:
    ontop_table_bounds = modify_pose_bounds_to_ be_ontop_of_object(init_state, env, init_bounds, banana.category, table.category)
    return position_within_bounds(banana.pose, ontop_table_bounds)
```

Here is another example set of functions generated for the task "serve the banana inside the blue thing after drying it by placing on the plate". The initial state for this example 
is shown in one of the attached images. Here, `banana` and `bowl` are both objects (the bowl happens to be blue, whereas the plate is red).
The initial state in this example is:
bowl: Pose=RavenPose(x=-0.09269248694181442, y=-0.7042829990386963, z=0.026169249787926674, roll=0.0, pitch=-0.0, yaw=0.8605557025412023)
banana: Pose=RavenPose(x=0.17416073374449514, y=-0.33348321026557554, z=0.02017684663429707, roll=5.081222700168695e-05, pitch=0.00013538346655467005, yaw=-3.0371082921616765)
plate: Pose=RavenPose(x=-0.11636300384998322, y=-0.4429782032966614, z=0.014744692512349077, roll=7.884650441866775e-28, pitch=-7.554679105908491e-28, yaw=2.245637386214381)
table: Pose=RavenPose(x=0.0, y=-0.5, z=0.0, roll=0.0, pitch=-0.0, yaw=0.0)

Importantly, notice how the `goal_check` function checks that the banana is 'upright' in the bowl by checking its rotation is 90 degrees (approx. 1.57 radians)
along the roll axis. This is necessary, because the banana only fits into the bowl in this orientation, as shown in another example rendered
image attached below. Pay careful attention and think about any similar orientation constraints that might be necessary in new problems. 

```python
def goal_check0() -> bool:
    in_bowl_bounds = modify_pose_bounds_to_ be_inside_object(init_state, env, init_bounds, bowl.category)
    banana_in_bowl_bounds = position_within_bounds(banana.pose, in_bowl_bounds)
    is_upright = 1.4 <= abs(banana.pose.roll) <= 1.65
    return banana_in_bowl_bounds and is_upright
```
Notice here that only one `goal_check` function was each defined, because satisfying the goal depends on all the continuous variables jointly.
Notice also that the `goal_check` doesn't test for anything to do with the plate, even though "drying" the banana in the plate was important to the task. This is because - in the final state - the banana should be in the bowl (it should have previously been placed in the cup), and the `goal_check` function only operates in the final state.

Finally, here's an example of constraints for a task "serve spam from its can into the cup". Here, the objects available are `potted_meat_can` and `mug`.
```python
def goal_check0() -> bool:
    above_mug_bounds = modify_pose_bounds_to_ be_above_object(init_state, env, init_bounds, mug.category)
    above_mug = position_within_ bounds(potted_meat_can.pose, above_bowl_bounds)
    pour_angle_sufficient = abs(potted_meat_can.pose.roll) > 1.2
    return above_bowl and pour_angle_sufficient
```
Notice once again that only one `goal_check` function was defined.
Notice also that the function checks the roll of the `potted_meat_can`, because this is important to know that it has been sufficiently 'tipped-over' such that its contents can fall from the bowl inside it into the cup.

Carefully consider these examples to inform your own functions for the current problem.
...
\end{lstlisting}

We use these same few-shot examples across all tasks without varying them.
We prompt the model to output \texttt{goal\_check} functions for the current goal and extract these.
Then, for each of the previous actions in the plan with language descriptions as well as predicates that rely on the VLM for their implementation, we prompt the VLM for continuous constraints that should hold after that particular action is executed.

\subsection{Code as Policies Full Prompt}
\label{appendix:cap-prompt}
Our full prompt for Code as Policies is shown below. Note that variables within curly brackets (\{\}) are filled-in dynamically depending on the task.

\begin{lstlisting}[style=promptstyle]
You are operating in an environment that has access to the following classes: {env_structs}.

You have access to the following variables. Note that TABLE_AABB.lower and TABLE_AABB.upper are tuples of length 3 that represent the lower and upper bounds on the x, y, and z positions that are on the table.
{typed_variables}

You also have access to the following objects in the environment. 
Each of these has type `RavenObject`. You are not able to use other objects:
{object_names}

Note that when you use/invoke these objects, do not use them as a string (i.e., do not provide quotes ''). Instead 
use them directly (e.g. use banana, not 'banana'). Use the name of the object as a string only where explicitly required
(e.g. by the GraspSampler).

Here's the poses of all the objects in the initial state of the scene (depicted in an attached image). Note that the axes of each object is shown as red (x-axis), green (y-axis), and blue (z-axis).
Note that roll is rotation about the x-axis, pitch is rotation about the y-axis and yaw is rotation about the z-axis. Pay careful attention to the axes and the current orientations of objects in the initial state as and when you decide to write functions involving orientations.
{init_state}

You also have access to helper functions whose signatures and docstrings are shown below. Pay careful attention to the arguments and return values of each function.
Helper function signatures:
{helper_functions_and_docstrings}

You have access to the following set of skills expressed as pddl predicates followed by descriptions. 
You have no other skills you can use, and you must exactly follow the number of inputs described below.
The coordinate axes are x, y, z where x is distance from the robot base, y is left/right from the robot base, and z is the height off the table.

Action("pick", [[o], g])
Pick up object o at grasp g sampled from a grasp sampler. Note that you should use the object o directly and not use its name (e.g. banana and not 'banana').

Action("place_ontop", [[o, s], p])
If holding an object o (e.g. `banana`) place the object ontop surface s (e.g. `table`) at pose p. Note that you should invoke s and o as objects directly and not use their names (e.g. banana and not 'banana').

Action("place_inside", [[o, c], p])
If holding an object o (e.g. `banana`) place the object inside container c (e.g. `bowl`) at placement pose p. Note that you should invoke o and c as objects directly and not use their names (e.g. banana and not 'banana').

Action("pour", [[o, c], p])
If holding an object `o` (e.g. `tomato_soup_can`) pour from the object onto/inside of container `c` (e.g. `bowl`) at a pouring pose dictated by p. Specifically, p is 4 numbers: <x, y, z> corresponding to the position to reach, and <theta> corresponding to the angle (in radians; -3.14 to 3.14) to tilt the hand to accomplish the pour. Note that you should invoke o and c as objects directly and not use their names (e.g. banana and not 'banana').

Finally, to help you generate continuous parameters required for these skills, you have access to the following samplers.

@dataclass
class GraspSampler(Sampler):
    curr_state: RavenState
    env: RavenYCBEnv
    object_name: str

    def sample(self, rng: np.random.Generator) -> List[float]:
        """Simply return some value within the AABB of the object, and at any
        orientation."""
        aabb_lb, aabb_ub = get_aabb_bounds(
            self.curr_state, self.env, self.object_name
        )
        lb = (aabb_lb[0], aabb_lb[1], aabb_lb[2], -3.14159265, -3.14159265, -3.14159265)
        ub = (aabb_ub[0], aabb_ub[1], aabb_ub[2], 3.14159265, 3.14159265, 3.14159265)
        return rng.uniform(lb, ub).tolist()

@dataclass
class PlaceSampler(Sampler):
    curr_state: RavenState
    env: RavenYCBEnv
    object_name_to_place_on_or_inside: str

    def sample(self, rng: np.random.Generator) -> List[float]:
        """Simply return some value within the AABB of the object, and at any
        orientation."""
        min_drop_height = 0.01
        max_drop_height = 0.35
        aabb_lb, aabb_ub = get_aabb_bounds(
            self.curr_state, self.env, 
            self.object_name_to_ place_on_or_inside
        )
        lb = (
            aabb_lb[0],
            aabb_lb[1],
            aabb_lb[2] + min_drop_height,
            -3.14159265,
            -3.14159265,
            -3.14159265,
        )
        ub = (
            aabb_ub[0],
            aabb_ub[1],
            aabb_ub[2] + max_drop_height,
            3.14159265,
            3.14159265,
            3.14159265,
        )
        return rng.uniform(lb, ub).tolist()

@dataclass
class PourSampler(Sampler):
    curr_state: RavenState
    env: RavenYCBEnv
    holding_obj_name: str
    obj_name_to_pour_into_or_ontop: str

    def sample(self, rng: np.random.Generator) -> List[float]:
        """Return a position sampled from above obj_name_to_pour_ into_or_ontop
        and a pitch randomly sampled between bounds."""
        # We sample (x, y, z, pitch.)
        obj_holding_lb, obj_holding_ub = get_aabb_bounds(
            self.curr_state, self.env, self.holding_obj_name
        )
        min_pour_height = obj_holding_ub[2] - obj_holding_lb[2]
        max_pour_height = min_pour_height * 2
        aabb_lb, aabb_ub = get_aabb_bounds(
            self.curr_state, self.env, self.obj_name_to_pour _into_or_ontop
        )
        pitch_range = (-3.14, 3.14)
        lb = (aabb_lb[0], aabb_lb[1], aabb_lb[2] + min_pour_height, pitch_range[0])
        ub = (aabb_ub[0], aabb_ub[1], aabb_ub[2] + max_pour_height, pitch_range[1])
        return rng.uniform(lb, ub).tolist()

Your goal is to generate a python function that returns a plan that performs the provided task. This function can
use helper functions that must be defined within the scope of the function itself.

The main function should be named EXACTLY `gen_plan`, and it should take in only one parameter corresponding to the environment state as input. Do not change the names. Do not create any additional classes or overwrite any existing ones. You are only allowed to create helper functions inside the `gen_plan` function.

Current provided goal: {provided_task_goal}

Here is an example `gen_plan` function for a different task:
#define user
Init state: 
bowl: Pose=RavenPose(x=-0.09269248694181442, y=-0.7042829990386963, z=0.026169249787926674, roll=0.0, pitch=-0.0, yaw=0.8605557025412023)
banana: Pose=RavenPose(x=0.17416073374449514, y=-0.33348321026557554, z=0.02017684663429707, roll=5.081222700168695e-05, pitch=0.00013538346655467005, yaw=-3.0371082921616765)
plate: Pose=RavenPose(x=-0.11636300384998322, y=-0.4429782032966614, z=0.014744692512349077, roll=7.884650441866775e-28, pitch=-7.554679105908491e-28, yaw=2.245637386214381)
table: Pose=RavenPose(x=0.0, y=-0.5, z=0.0, roll=0.0, pitch=-0.0, yaw=0.0)
Task goal: put the lemon on the plate and the banana on the table

#define assistant
```python
def gen_plan(initial:RavenState):
    plan = []
    place_pose = PlaceSampler(initial, env, "plate").sample(rng)
    lemon_grasp = GraspSampler(initial, env, "lemon").sample(rng)
    plan.append(Action("pick", [[lemon], lemon_grasp]))
    plan.append(Action("place_ontop", [[lemon, plate], place_pose]))
    place_pose = PlaceSampler(initial, env, "table").sample(rng)
    banana_grasp = GraspSampler(initial, env, "banana").sample(rng)
    plan.append(Action("pick", [[banana], banana_grasp]))
    plan.append(Action("place_ontop", [[banana, table], RavenPose(x=x_place, y=y_place, z=0.02)]))
    return plan
```

Here is another example for a different goal involving the same objects and initial state. Note that in this case, the bowl is blue.
Also note that the banana needs to be rotated so that it fits into the bowl; in general you should pay careful 
attention to any angular constraints that might be important for solving different tasks.

#define user
Init state: 
bowl: Pose=RavenPose(x=-0.09269248694181442, y=-0.7042829990386963, z=0.026169249787926674, roll=0.0, pitch=-0.0, yaw=0.8605557025412023)
banana: Pose=RavenPose(x=0.17416073374449514, y=-0.33348321026557554, z=0.02017684663429707, roll=5.081222700168695e-05, pitch=0.00013538346655467005, yaw=-3.0371082921616765)
plate: Pose=RavenPose(x=-0.11636300384998322, y=-0.4429782032966614, z=0.014744692512349077, roll=7.884650441866775e-28, pitch=-7.554679105908491e-28, yaw=2.245637386214381)
table: Pose=RavenPose(x=0.0, y=-0.5, z=0.0, roll=0.0, pitch=-0.0, yaw=0.0)
Task goal: serve the banana inside the blue thing after drying it by placing on the plate

#define assistant
```python
def gen_plan(initial:RavenState):
    plan = []
    place_pose = PlaceSampler(initial, env, "plate").sample(rng)
    banana_grasp = GraspSampler(initial, env, "banana").sample(rng)
    plan.append(Action("pick", [[banana], grasp]))
    plan.append(Action("place_ontop", [[banana, plate], place_pose]))
    bowl_bounds = get_aabb_bounds(initial, env, "bowl")
    place_pose = PlaceSampler(initial, env, "bowl").sample(rng)
    bowl_center_pose = get_obj_center(initial, env)
    # Drop the object in the exact center of the bowl
    place_pose[0] = bowl_center_pose[0]
    place_pose[1] = bowl_center_pose[1]
    place_pose[2] = bowl_center_pose + 0.05
    # Make sure to pick an orientation such that the banana will fit!
    place_pose.roll = 1.5
    banana_grasp = GraspSampler(initial, env, "banana").sample(rng)
    plan.append(Action("pick", [[banana], grasp]))
    plan.append(Action("place_inside", [[banana, bowl], place_pose]))
    return plan
```

Here is yet another example for a different task:
mug: Pose=RavenPose(x=-0.09269248694181442, y=-0.7042829990386963, z=0.026169249787926674, roll=0.0, pitch=-0.0, yaw=0.8605557025412023)
potted_meat_can: Pose=RavenPose(x=0.17416073374449514, y=-0.33348321026557554, z=0.02017684663429707, roll=5.081222700168695e-05, pitch=0.00013538346655467005, yaw=-3.0371082921616765)
plate: Pose=RavenPose(x=-0.11636300384998322, y=-0.4429782032966614, z=0.014744692512349077, roll=7.884650441866775e-28, pitch=-7.554679105908491e-28, yaw=2.245637386214381)
table: Pose=RavenPose(x=0.0, y=-0.5, z=0.0, roll=0.0, pitch=-0.0, yaw=0.0)
Task goal: serve spam from its can into the cup

#define assistant
```python
def gen_plan(initial:RavenState):
    plan = []
    spam_grasp = GraspSampler(initial, env, "potted_meat_can").sample(rng)
    pour_params = PourSampler(initial, env, "potted_meat_can", "cup").sample(rng)
    pour_params[3] = 2.0  # necessary for the contents to really fall into the cup.
    plan.append(Action("pick", [[potted_meat_can], spam_grasp]))
    plan.append(Action("pour", [[potted_meat_can, mug], pour_params]))
    return plan
```

Make sure to enclose your output  with ```python <output gen_plan(initial:Ravenstate): function> ``` (ignore the angle brackets - those are just for illustrative purposes).
\end{lstlisting}

%% file: references.bib
@misc{black2024pi0visionlanguageactionflowmodel,
      title={$\pi_0$: A Vision-Language-Action Flow Model for General Robot Control}, 
      author={Kevin Black and Noah Brown and Danny Driess and Adnan Esmail and Michael Equi and Chelsea Finn and Niccolo Fusai and Lachy Groom and Karol Hausman and Brian Ichter and Szymon Jakubczak and Tim Jones and Liyiming Ke and Sergey Levine and Adrian Li-Bell and Mohith Mothukuri and Suraj Nair and Karl Pertsch and Lucy Xiaoyang Shi and James Tanner and Quan Vuong and Anna Walling and Haohuan Wang and Ury Zhilinsky},
      year={2024},
      eprint={2410.24164},
      archivePrefix={arXiv},
      primaryClass={cs.LG},
      url={https://arxiv.org/abs/2410.24164}, 
}

@misc{nvidia2025gr00tn1openfoundation,
      title={GR00T N1: An Open Foundation Model for Generalist Humanoid Robots}, 
      author={NVIDIA and : and Johan Bjorck and Fernando Castañeda and Nikita Cherniadev and Xingye Da and Runyu Ding and Linxi "Jim" Fan and Yu Fang and Dieter Fox and Fengyuan Hu and Spencer Huang and Joel Jang and Zhenyu Jiang and Jan Kautz and Kaushil Kundalia and Lawrence Lao and Zhiqi Li and Zongyu Lin and Kevin Lin and Guilin Liu and Edith Llontop and Loic Magne and Ajay Mandlekar and Avnish Narayan and Soroush Nasiriany and Scott Reed and You Liang Tan and Guanzhi Wang and Zu Wang and Jing Wang and Qi Wang and Jiannan Xiang and Yuqi Xie and Yinzhen Xu and Zhenjia Xu and Seonghyeon Ye and Zhiding Yu and Ao Zhang and Hao Zhang and Yizhou Zhao and Ruijie Zheng and Yuke Zhu},
      year={2025},
      eprint={2503.14734},
      archivePrefix={arXiv},
      primaryClass={cs.RO},
      url={https://arxiv.org/abs/2503.14734}, 
}

@article{wei2023chainofthoughtpromptingelicitsreasoning,
  title={Chain-of-thought prompting elicits reasoning in large language models},
  author={Wei, Jason and Wang, Xuezhi and Schuurmans, Dale and Bosma, Maarten and Xia, Fei and Chi, Ed and Le, Quoc V and Zhou, Denny and others},
  journal={Advances in neural information processing systems},
  volume={35},
  pages={24824--24837},
  year={2022},
  url={https://proceedings.neurips.cc/paper_files/paper/2022/hash/9d5609613524ecf4f15af0f7b31abca4-Abstract-Conference.html},
}

@inproceedings{guo2024castlconstraintsspecificationsllm,
      title={CaStL: Constraints as Specifications through LLM Translation for Long-Horizon Task and Motion Planning}, 
      author={Weihang Guo and Zachary Kingston and Lydia E. Kavraki},
      year={2024},
      eprint={2410.22225},
      archivePrefix={arXiv},
      primaryClass={cs.RO},
      booktitle={arxiv preprint},
      url={https://arxiv.org/abs/2410.22225}, 
}

@inproceedings{athalye2024predicate,
  title={Predicate Invention from Pixels via Pretrained Vision-Language Models},
  author={Athalye, Ashay and Kumar, Nishanth and Silver, Tom and Liang, Yichao and Lozano-P{\'e}rez, Tom{\'a}s and Kaelbling, Leslie Pack},
  booktitle={arXiv preprint arXiv:2501.00296},
  url={https://arxiv.org/pdf/2501.00296},
  year={2024}
}

@inproceedings{liang2024visualpredicatorlearningabstractworld,
      title={VisualPredicator: Learning Abstract World Models with Neuro-Symbolic Predicates for Robot Planning},
      author={Yichao Liang and Nishanth Kumar and Hao Tang and Adrian Weller and Joshua B. Tenenbaum and Tom Silver and João F. Henriques and Kevin Ellis},
      year={2024},
      booktitle={International Conference on Learning Representations},
      url={https://arxiv.org/abs/2410.23156},
}

@misc{ma2024eurekahumanlevelrewarddesign,
    title={Eureka: Human-Level Reward Design via Coding Large Language Models}, 
    author={Yecheng Jason Ma and William Liang and Guanzhi Wang and De-An Huang and Osbert Bastani and Dinesh Jayaraman and Yuke Zhu and Linxi Fan and Anima Anandkumar},
    year={2024},
    eprint={2310.12931},
    archivePrefix={arXiv},
    primaryClass={cs.RO},
    url={https://arxiv.org/abs/2310.12931}, 
}

@inproceedings{xie2023translating,
    title={Translating Natural Language to Planning Goals with Large-Language Models}, 
    author={Yaqi Xie and Chen Yu and Tongyao Zhu and Jinbin Bai and Ze Gong and Harold Soh},
    year={2023},
    eprint={2302.05128},
    archivePrefix={arXiv},
    primaryClass={cs.CL},
    booktitle={arxiv preprint},
    url={https://arxiv.org/abs/2302.05128}, 
}

@misc{hu2023lookleapunveilingpower,
    title={Look Before You Leap: Unveiling the Power of GPT-4V in Robotic Vision-Language Planning}, 
    author={Yingdong Hu and Fanqi Lin and Tong Zhang and Li Yi and Yang Gao},
    year={2023},
    eprint={2311.17842},
    archivePrefix={arXiv},
    primaryClass={cs.RO},
    url={https://arxiv.org/abs/2311.17842}, 
}

@inproceedings{bonet1999planning,
    title={Planning as heuristic search: New results},
    author={Bonet, Blai and Geffner, Hector},
    booktitle={European Conference on Planning},
    year={1999},
    organization={Springer},
    url={https://bonetblai.github.io/reports/ecp99-hspr.pdf}
}

@article{bonet2001planning,
    title = {{Planning as heuristic search}},
    year = {2001},
    journal = {Artificial Intelligence},
    author = {Bonet, Blai and Geffner, Héctor},
    url={https://www.cs.toronto.edu/~sheila/2542/s14/A1/bonetgeffner-heusearch-aij01.pdf}
}

@article{HoffmannN01,
    title = {{The {\{}FF{\}} Planning System: Fast Plan Generation Through Heuristic Search}},
    year = {2001},
    journal = {Journal Artificial Intelligence Research (JAIR)},
    author = {Hoffmann, Jörg and Nebel, Bernhard},
    pages = {253--302},
    volume = {14},
    url = {http://dl.acm.org/citation.cfm?id=1622404}
}

@inproceedings{edelkamp2000exhibiting,
    title={Exhibiting knowledge in planning problems to minimize state encoding length},
    author={Edelkamp, Stefan and Helmert, Malte},
    booktitle={Recent Advances in AI Planning: 5th European Conference on Planning},
    year={2000},
    organization={Springer},
    url={https://ai.dmi.unibas.ch/papers/edelkamp-helmert-ecp1999.pdf}
}

@article{helmert2006fast,
    title = {{The Fast Downward Planning System}},
    year = {2006},
    journal = {Journal of Artificial Intelligence Research (JAIR)},
    author = {Helmert, Malte},
    pages = {191--246},
    volume = {26},
    url = {http://www.jair.org/papers/paper1705.html}
}

@techreport{mcdermott1998pddl,
    author = {McDermott, Drew and Ghallab, Malik and Howe, Adele and Knoblock, Craig and Ram, Ashwin and Veloso, Manuela and Weld, Daniel and Wilkins, David},
    institution = {Yale Center for Computational Vision and Control},
    title = {{PDDL: The Planning Domain Definition Language}},
    year = {1998},
    url = {https://www.cs.cmu.edu/~mmv/planning/readings/98aips-PDDL.pdf}
}

@article{garrett2018stripstream,
  title={STRIPStream: Integrating Symbolic Planners and Blackbox Samplers},
  author={Caelan Reed Garrett and Tomas Lozano-Perez and Leslie Pack Kaelbling},
  booktitle={arxiv preprint},
  year={2018},
  volume={abs/1802.08705},
  url={https://export.arxiv.org/pdf/1802.08705}
}

@article{garrett2018sampling,
    title={Sampling-based methods for factored task and motion planning},
    author={Garrett, Caelan Reed and Lozano-P{\'e}rez, Tom{\'a}s and Kaelbling, Leslie Pack},
    journal={International Journal of Robotics Research (IJRR)},
    year={2018},
    publisher={SAGE Publications Sage UK: London, England},
    url={https://arxiv.org/pdf/1801.00680}
}

@inproceedings{garrett2020PDDLStream,
    title = {{PDDLStream: Integrating Symbolic Planners and Blackbox Samplers}},
    year = {2020},
    booktitle = {International Conference on Automated Planning and Scheduling (ICAPS)},
    author = {Garrett, Caelan R. and Lozano-P{\'{e}}rez, Tomás and Kaelbling, Leslie P.},
    url = {https://arxiv.org/abs/1802.08705}
}

@phdthesis{garrett2021sampling,
    title={Sampling-Based Robot Task and Motion Planning in the Real World},
    author={Garrett, Caelan Reed},
    year={2021},
    school={Massachusetts Institute of Technology},
    url={https://dspace.mit.edu/bitstream/handle/1721.1/139990/Garrett-cgarrett-PhD-EECS-2021-thesis.pdf?sequence=1\&isAllowed=y}
}

@article{kambhampati2024can,
    title={Can large language models reason and plan?},
    author={Kambhampati, Subbarao},
    journal={Annals of the New York Academy of Sciences},
    year={2024},
    publisher={Wiley Online Library},
    url={https://arxiv.org/pdf/2403.04121}
}

@inproceedings{
    curtis2024trustproc3ssolvinglonghorizon,
    title={Trust the {PR}oC3S: Solving Long-Horizon Robotics Problems with {LLM}s and Constraint Satisfaction},
    author={Aidan Curtis and Nishanth Kumar and Jing Cao and Tom{\'a}s Lozano-P{\'e}rez and Leslie Pack Kaelbling},
    booktitle={Conference on Robot Learning (CoRL)},
    year={2024},
    url={https://openreview.net/forum?id=r6ZhiVYriY}
}

@inproceedings{
    huang2024rekep,
    title={ReKep: Spatio-Temporal Reasoning of Relational Keypoint Constraints for Robotic Manipulation},
    author={Wenlong Huang and Chen Wang and Yunzhu Li and Ruohan Zhang and Li Fei-Fei},
    booktitle={Conference on Robot Learning (CoRL)},
    year={2024},
    url={https://openreview.net/forum?id=9iG3SEbMnL}
}

@inproceedings{
    yuan2024robopoint,
    title={RoboPoint: A Vision-Language Model for Spatial Affordance Prediction in Robotics},
    author={Wentao Yuan and Jiafei Duan and Valts Blukis and Wilbert Pumacay and Ranjay Krishna and Adithyavairavan Murali and Arsalan Mousavian and Dieter Fox},
    booktitle={Conference on Robot Learning (CoRL)},
    year={2024},
    url={https://openreview.net/forum?id=GVX6jpZOhU}
}

@inproceedings{
    duan2024manipulateanything,
    title={Manipulate-Anything: Automating Real-World Robots using Vision-Language Models},
    author={Jiafei Duan and Wentao Yuan and Wilbert Pumacay and Yi Ru Wang and Kiana Ehsani and Dieter Fox and Ranjay Krishna},
    booktitle={Conference on Robot Learning (CoRL)},
    year={2024},
    url={https://openreview.net/forum?id=2SYFDG4WRA}
}

@article{llm_plus_p,
    title={LLM+P: Empowering Large Language Models with Optimal Planning Proficiency},
    author={B. Liu and Yuqian Jiang and Xiaohan Zhang and Qian Liu and Shiqi Zhang and Joydeep Biswas and Peter Stone},
    journal={ArXiv},
    year={2023},
    volume={abs/2304.11477},
    url={https://api.semanticscholar.org/CorpusID:258298051}
}

@misc{yang2024guidinglonghorizontaskmotion,
      title={Guiding Long-Horizon Task and Motion Planning with Vision Language Models}, 
      author={Zhutian Yang and Caelan Garrett and Dieter Fox and Tomás Lozano-Pérez and Leslie Pack Kaelbling},
      year={2024},
      eprint={2410.02193},
      archivePrefix={arXiv},
      primaryClass={cs.RO},
      url={https://arxiv.org/abs/2410.02193}, 
}

@inproceedings{huang2023voxposer,
    title={VoxPoser: Composable 3D Value Maps for Robotic Manipulation with Language Models}, 
    author={Wenlong Huang and Chen Wang and Ruohan Zhang and Yunzhu Li and Jiajun Wu and Li Fei-Fei},
    year={2023},
    booktitle={Conference on Robot Learning (CoRL)},
    url={https://voxposer.github.io/voxposer.pdf}
}

@inproceedings{inner_monologue,
    title={Inner Monologue: Embodied Reasoning through Planning with Language Models}, 
    author={Wenlong Huang and Fei Xia and Ted Xiao and Harris Chan and Jacky Liang and Pete Florence and Andy Zeng and Jonathan Tompson and Igor Mordatch and Yevgen Chebotar and Pierre Sermanet and Noah Brown and Tomas Jackson and Linda Luu and Sergey Levine and Karol Hausman and Brian Ichter},
    year={2023},
    booktitle={Conference on Robot Learning (CoRL)},
    url={https://openreview.net/pdf?id=3R3Pz5i0tye}
}

@inproceedings{tamp_llm,
    title={Task and motion planning with large language models for object rearrangement},
    author={Ding, Yan and Zhang, Xiaohan and Paxton, Chris and Zhang, Shiqi},
    booktitle={2023 IEEE/RSJ International Conference on Intelligent Robots and Systems (IROS)},
    year={2023},
    url={https://arxiv.org/pdf/2303.06247}
}

@InProceedings{saycan,
    title = 	 {Do As I Can, Not As I Say: Grounding Language in Robotic Affordances},
    author =       {Ichter, Brian and Brohan, Anthony and Chebotar, Yevgen and Finn, Chelsea and Hausman, Karol and Herzog, Alexander and Ho, Daniel and Ibarz, Julian and Irpan, Alex and Jang, Eric and Julian, Ryan and Kalashnikov, Dmitry and Levine, Sergey and Lu, Yao and Parada, Carolina and Rao, Kanishka and Sermanet, Pierre and Toshev, Alexander T and Vanhoucke, Vincent and Xia, Fei and Xiao, Ted and Xu, Peng and Yan, Mengyuan and Brown, Noah and Ahn, Michael and Cortes, Omar and Sievers, Nicolas and Tan, Clayton and Xu, Sichun and Reyes, Diego and Rettinghouse, Jarek and Quiambao, Jornell and Pastor, Peter and Luu, Linda and Lee, Kuang-Huei and Kuang, Yuheng and Jesmonth, Sally and Joshi, Nikhil J. and Jeffrey, Kyle and Ruano, Rosario Jauregui and Hsu, Jasmine and Gopalakrishnan, Keerthana and David, Byron and Zeng, Andy and Fu, Chuyuan Kelly},
    booktitle = 	 {Conference on Robot Learning (CoRL)},
    year = 	 {2023},
    url = 	 {https://proceedings.mlr.press/v205/ichter23a.html},
}

@inproceedings{llm3,
    title={LLM$^3$:Large Language Model-based Task and Motion Planning with Motion Failure Reasoning}, 
    author={Shu Wang and Muzhi Han and Ziyuan Jiao and Zeyu Zhang and Ying Nian Wu and Song-Chun Zhu and Hangxin Liu},
    year={2024},
    booktitle={arXiv preprint},
    url={https://arxiv.org/pdf/2403.11552}
}

@inproceedings{code_as_policy,
    title={Code as Policies: Language Model Programs for Embodied Control},
    author={Jacky Liang and Wenlong Huang and Fei Xia and Peng Xu and Karol Hausman and Brian Ichter and Pete Florence and Andy Zeng},
    booktitle={arXiv preprint},
    year={2022},
    url={https://arxiv.org/pdf/2209.07753}
}

@inproceedings{prog_prompt,
    title={Progprompt: Generating situated robot task plans using large language models},
    author={Singh, Ishika and Blukis, Valts and Mousavian, Arsalan and Goyal, Ankit and Xu, Danfei and Tremblay, Jonathan and Fox, Dieter and Thomason, Jesse and Garg, Animesh},
    booktitle={IEEE International Conference on Robotics and Automation (ICRA)},
    year={2023},
    url={https://arxiv.org/pdf/2209.11302}
}

@inproceedings{grounded_decoding,
    title={Grounded Decoding: Guiding Text Generation with Grounded Models for Embodied Agents}, 
    author={Wenlong Huang and Fei Xia and Dhruv Shah and Danny Driess and Andy Zeng and Yao Lu and Pete Florence and Igor Mordatch and Sergey Levine and Karol Hausman and Brian Ichter},
    year={2023},
    eprint={2303.00855},
    archivePrefix={arXiv},
    booktitle={arxiv preprint},
    primaryClass={cs.RO},
    url={https://arxiv.org/pdf/2303.00855}
}

@inproceedings{doremi,
    title={DoReMi: Grounding Language Model by Detecting and Recovering from Plan-Execution Misalignment}, 
    author={Yanjiang Guo and Yen-Jen Wang and Lihan Zha and Zheyuan Jiang and Jianyu Chen},
    year={2023},
    eprint={2307.00329},
    archivePrefix={arXiv},
    primaryClass={cs.RO},
    booktitle={arxiv preprint},
    url={https://arxiv.org/pdf/2307.00329}
}

@inproceedings{replan,
    title={RePLan: Robotic Replanning with Perception and Language Models}, 
    author={Marta Skreta and Zihan Zhou and Jia Lin Yuan and Kourosh Darvish and Alán Aspuru-Guzik and Animesh Garg},
    year={2024},
    eprint={2401.04157},
    archivePrefix={arXiv},
    primaryClass={cs.RO},
    booktitle={arxiv preprint},
    url={https://arxiv.org/pdf/2401.04157}
}

@inproceedings{chen2024autotamp,
    title={AutoTAMP: Autoregressive Task and Motion Planning with LLMs as Translators and Checkers}, 
    author={Yongchao Chen and Jacob Arkin and Charles Dawson and Yang Zhang and Nicholas Roy and Chuchu Fan},
    year={2024},
    eprint={2306.06531},
    archivePrefix={arXiv},
    booktitle={arXiv preprint},
    primaryClass={cs.RO},
    url={https://arxiv.org/pdf/2306.06531}
}

@article{tamp_survey,
    author       = {Caelan Reed Garrett and
                  Rohan Chitnis and
                  Rachel M. Holladay and
                  Beomjoon Kim and
                  Tom Silver and
                  Leslie Pack Kaelbling and
                  Tom{\'{a}}s Lozano{-}P{\'{e}}rez},
    title        = {Integrated Task and Motion Planning},
    journal      = {CoRR},
    volume       = {abs/2010.01083},
    year         = {2020},
    url          = {https://arxiv.org/abs/2010.01083},
    eprinttype    = {arXiv},
    eprint       = {2010.01083},
    bibsource    = {dblp computer science bibliography, https://dblp.org}
}

@inproceedings{garrett2020online,
    title = {{Online Replanning in Belief Space for Partially Observable Task and Motion Problems}},
    year = {2020},
    booktitle = {IEEE International Conference on Robotics and Automation (ICRA)},
    author = {Garrett, Caelan R. and Paxton, Chris and Lozano-P{\'{e}}rez, Tomás and Kaelbling, Leslie P. and Fox, Dieter},
    url = {https://arxiv.org/abs/1911.04577}
}

@inproceedings{m0m,
    title={Long-horizon manipulation of unknown objects via task and motion planning with estimated affordances},
    author={Curtis, Aidan and Fang, Xiaolin and Kaelbling, Leslie Pack and Lozano-P{\'e}rez, Tom{\'a}s and Garrett, Caelan Reed},
    booktitle={IEEE International Conference on Robotics and Automation (ICRA)},
    year={2022},
    url={https://arxiv.org/pdf/2108.04145}
}

@article{bilevel-planning-blog-post,
    author    = {Nishanth Kumar and Willie McClinton and Kathryn Le and Tom Silver},
    title     = {Bilevel Planning for Robots: An Illustrated Introduction},
    url      = {https://lis.csail.mit.edu/bilevel-planning-for-robots-an-illustrated-introduction},
    year      = {2023}
}

@article{zeng2020transporter,
    title={Transporter Networks: Rearranging the Visual World for Robotic Manipulation},
    author={Zeng, Andy and Florence, Pete and Tompson, Jonathan and Welker, Stefan and Chien, Jonathan and Attarian, Maria and Armstrong, Travis and Krasin, Ivan and Duong, Dan and Sindhwani, Vikas and Lee, Johnny},
    journal={Conference on Robot Learning (CoRL)},
    year={2020},
    url={https://proceedings.mlr.press/v155/zeng21a/zeng21a.pdf}
}

@inproceedings{silver2023predicate,
    title={Predicate invention for bilevel planning},
    author={Silver, Tom and Chitnis, Rohan and Kumar, Nishanth and McClinton, Willie and Lozano-P{\'e}rez, Tom{\'a}s and Kaelbling, Leslie and Tenenbaum, Joshua B},
    booktitle={AAAI Conference on Artificial Intelligence (AAAI)},
    year={2023},
    url={https://ojs.aaai.org/index.php/AAAI/article/view/26429/26201}
}

@inproceedings{
    kumar2023learning,
    title={Learning Efficient Abstract Planning Models that Choose What to Predict},
    author={Nishanth Kumar and Willie McClinton and Rohan Chitnis and Tom Silver and Tom{\'a}s Lozano-P{\'e}rez and Leslie Pack Kaelbling},
    booktitle={Conference on Robot Learning (CoRL)},
    year={2023},
    url={https://openreview.net/pdf?id=\_gZLyRGGuo}
}

@article{konidaris2018skills,
    title={From skills to symbols: Learning symbolic representations for abstract high-level planning},
    author={Konidaris, George and Kaelbling, Leslie Pack and Lozano-P{\'e}rez, Tom{\'a}s},
    journal={Journal of Artificial Intelligence Research (JAIR)},
    year={2018},
    url={https://jair.org/index.php/jair/article/view/11175/26380}
}

@inproceedings{srivastava2014combined,
    title={Combined task and motion planning through an extensible planner-independent interface layer},
    author={Srivastava, Siddharth and Fang, Eugene and Riano, Lorenzo and Chitnis, Rohan and Russell, Stuart and Abbeel, Pieter},
    booktitle={IEEE international conference on robotics and automation (ICRA)},
    year={2014},
    url={https://people.eecs.berkeley.edu/~russell/papers/icra14-planrob.pdf}
}

@inproceedings{huang2024copa,
    title={CoPa: General Robotic Manipulation through Spatial Constraints of Parts with Foundation Models}, 
    author={Haoxu Huang and Fanqi Lin and Yingdong Hu and Shengjie Wang and Yang Gao},
    year={2024},
    eprint={2403.08248},
    archivePrefix={arXiv},
    primaryClass={cs.RO},
    booktitle={arxiv preprint},
    url={https://arxiv.org/pdf/2403.08248}
}

@inproceedings{beeson2015trac,
  title={TRAC-IK: An open-source library for improved solving of generic inverse kinematics},
  author={Beeson, Patrick and Ames, Barrett},
  booktitle={IEEE-RAS 15th International Conference on Humanoid Robots (Humanoids)},
  year={2015},
  organization={IEEE},
  url={https://ieeexplore.ieee.org/stamp/stamp.jsp?tp=\&arnumber=7363472}
}

@inproceedings{sundaralingam2023curobo,
  title={Curobo: Parallelized collision-free robot motion generation},
  author={Sundaralingam, Balakumar and Hari, Siva Kumar Sastry and Fishman, Adam and Garrett, Caelan and Van Wyk, Karl and Blukis, Valts and Millane, Alexander and Oleynikova, Helen and Handa, Ankur and Ramos, Fabio and others},
  booktitle={IEEE International Conference on Robotics and Automation (ICRA)},
  year={2023},
  url={https://arxiv.org/pdf/2310.17274}
}

@inproceedings{goyal2023rvt,
  title={Rvt: Robotic view transformer for 3d object manipulation},
  author={Goyal, Ankit and Xu, Jie and Guo, Yijie and Blukis, Valts and Chao, Yu-Wei and Fox, Dieter},
  booktitle={Conference on Robot Learning (CoRL)},
  year={2023},
  url={https://arxiv.org/pdf/2306.14896}
}

@article{liu2023grounding,
  title={Grounding dino: Marrying dino with grounded pre-training for open-set object detection},
  author={Liu, Shilong and Zeng, Zhaoyang and Ren, Tianhe and Li, Feng and Zhang, Hao and Yang, Jie and Li, Chunyuan and Yang, Jianwei and Su, Hang and Zhu, Jun and others},
  journal={arXiv preprint arXiv:2303.05499},
  year={2023},
  url={https://arxiv.org/pdf/2303.05499}
}

@inproceedings{AgiaMigimatsuEtAl2023,
    title     = {STAP: Sequencing Task-Agnostic Policies}, 
    author    = {Agia, Christopher and Migimatsu, Toki and Wu, Jiajun and Bohg, Jeannette},
    booktitle = {2023 IEEE International Conference on Robotics and Automation (ICRA)}, 
    year      = {2023},
    url       = {https://arxiv.org/pdf/2210.12250}
}

@inproceedings{openai2024gpt4,
    title={GPT-4 Technical Report},
    author={OpenAI Josh Achiam and Steven Adler and Sandhini Agarwal and Lama Ahmad and Ilge Akkaya and Florencia Leoni Aleman and Diogo Almeida and Janko Altenschmidt and Sam Altman et. al.},
    booktitle={arxiv preprint},
    year={2023},
    url={https://arxiv.org/pdf/2303.08774}
}
